\documentclass[fleqn,10pt]{wlscirep}
\usepackage[utf8]{inputenc}
\usepackage[T1]{fontenc}
\usepackage{multirow}
\usepackage{amsmath,amssymb,amsfonts}
\usepackage{xcolor}
\usepackage{graphicx}
\usepackage{textcomp}

\title{Natural scene reconstruction from fMRI signals using generative latent diffusion}

\author[1,2,*]{Furkan Ozcelik}
\author[1,2,3]{Rufin VanRullen}
\affil[1]{CerCo, CNRS UMR5549, Toulouse, France}
\affil[2]{Universite de Toulouse, Toulouse, France}
\affil[3]{ANITI, Toulouse, France}
\affil[*]{ozcelikfu@gmail.com}

\begin{abstract}
In neural decoding research, one of the most intriguing topics is the reconstruction of perceived natural images based on fMRI signals. Previous studies have succeeded in re-creating different aspects of the visuals, such as low-level properties (shape, texture, layout) or high-level features (category of objects, descriptive semantics of scenes) but have typically failed to reconstruct these properties together for complex scene images. Generative AI has recently made a leap forward with latent diffusion models capable of generating high-complexity images. Here, we investigate how to take advantage of this innovative technology for brain decoding. We present a two-stage scene reconstruction framework called ``Brain-Diffuser''. In the first stage, starting from fMRI signals, we reconstruct images that capture low-level properties and overall layout using a VDVAE (Very Deep Variational Autoencoder) model. In the second stage, we use the image-to-image framework of a latent diffusion model (Versatile Diffusion) conditioned on predicted multimodal (text and visual) features, to generate final reconstructed images. On the publicly available Natural Scenes Dataset benchmark, our method outperforms previous models both qualitatively and quantitatively. When applied to synthetic fMRI patterns generated from individual ROI (region-of-interest) masks, our trained model creates compelling ``ROI-optimal'' scenes consistent with neuroscientific knowledge. Thus, the proposed methodology can have an impact on both applied (e.g. brain-computer interface) and fundamental neuroscience.

\end{abstract}
\begin{document}

\flushbottom
\maketitle
%
%
\thispagestyle{empty}


\section{Introduction}

 Establishing neural encoding and decoding techniques is one way for researchers to discover how the brain and cognition work. Recent developments in modeling and computation have opened up new ways of decoding information from brain signals. Numerous studies in the field of vision research have employed statistical techniques and machine learning to decode specific information from fMRI (functional Magnetic Resonance Imaging) neural activity, such as position~\cite{thirion2006inverse} or orientation~\cite{kamitani2005decoding,haynes2005predicting}, to predict categories of images~\cite{haxby2001distributed,cox2003functional}, to match exemplar images from a candidate set~\cite{kay2008identifying}, and to reconstruct images with low levels of complexity, such as simple shapes and structures~\cite{miyawaki2008visual}. 

In recent years, following the success in the development of deep learning models, many studies utilized deep generative models to reconstruct entire images. These deep generative models included Variational Autoencoders (VAE), Generative Adversarial Networks (GAN), and recently Latent Diffusion Models (LDM). Most of these studies used existing deep generative models, pretrained on large-scale data, and then learned a mapping (with simple regression or more advanced neural network architectures) to reconstruct the corresponding latent variables from the brain signals. This general method was used to reconstruct images with different levels of complexity such as faces~\cite{vanrullen2019reconstructing,dado2022hyperrealistic}, single-object-centered images~\cite{shen2019deep}, and more complex scenes~\cite{allen2022massive,lin2022mind}.

Most of the earlier works on natural scene reconstruction studied either the Generic Object Decoding~\cite{horikawa2017generic} or the Deep Image Reconstruction~\cite{shen2019deep} datasets curated by the Kamitani Lab. These datasets consist of 1200 training and 50 testing images from the ImageNet~\cite{deng2009imagenet} dataset and they differ in the number of fMRI repetitions for training and testing images. One of the pioneer studies in this area is by Shen et al.~\cite{shen2019deep} who optimized input images using a deep generator network with a loss function provided by fMRI-decoded CNN features. Beliy et al.~\cite{beliy2019voxels} utilized supervised training with \{fMRI, stimulus\} pairs, alongside an additional consistency loss for unsupervised training with test fMRI data and additional image data. Building on this, Gaziv et al.~\cite{gaziv2020self} further improved the method by incorporating a perceptual loss on reconstructed images, resulting in sharper reconstructions. Mozafari et al.~\cite{mozafari2020reconstructing} introduced a reconstruction model based on BigBiGAN that focused on semantics. Ren et al.~\cite{ren2021reconstructing} devised a dual VAE-GAN model with a three-stage learning strategy that incorporates adversarial learning and knowledge distillation. Ozcelik et al.~\cite{ozcelik2022reconstruction} employed the Instance-Conditioned GAN model to generate reconstructions focused on accurate semantics (by extracting semantic information with the SwAV model) and pose information (with latent optimization). Chen et al.~\cite{chen2023seeing} utilized a sparse masked brain modeling on large-scale fMRI data and then trained a double-conditioned diffusion model for visual decoding. 

Recently, Allen et al. curated another dataset for visual encoding and decoding studies called Natural Scenes Dataset~\cite{allen2022massive}. For this dataset, 8 subjects viewed thousands of images from the COCO~\cite{lin2014microsoft} dataset. COCO images contain multiple objects and they are more complex in nature compared to ImageNet images. Because of the number, diversity, and complexity of images included, the NSD dataset---although very recent---is becoming the de facto benchmark for fMRI-based natural scene reconstruction. Thus, it is the dataset that we chose for the present work. There are already three studies that reconstructed images from this dataset, and we can use them as baselines against which to compare our model's performance. The first one is by Lin et al.~\cite{lin2022mind}, who utilized the Lafite framework that adapts the StyleGAN2 model for text-to-image generation. Takagi et al.~\cite{takagi2023high} devised a method based on Stable Diffusion, using captions for the semantic information and latent variables from images for the low-level information. Gu et al.~\cite{gu2023decoding} improved upon Ozcelik et al.'s~\cite{ozcelik2022reconstruction} IC-GAN framework, by establishing a surface-based convolutional network to process fMRI data instead of using vectorized data in the regression models; they also trained an encoder network to predict pose information, instead of performing latent optimization. 

The above studies have fostered advances in reconstructing images with high fidelity, especially in the case of object-centered images (i.e., ImageNet images from the Kamitani dataset). Yet, reconstructing scenes with multiple objects and complex semantic descriptions (i.e., COCO images from the NSD dataset) remains a challenge. Given the remarkable recent success of latent diffusion models~\cite{rombach2022high}  in generative AI applications such as text-to-image generation~\cite{rombach2022high,ramesh2022hierarchical,nichol2022glide,saharia2022photorealistic,xu2022versatile}, we reasoned that brain decoding studies could also take advantage of such models.
Thus, we propose here a visual reconstruction framework called "Brain-Diffuser", relying on the powerful generation capabilities of Versatile Diffusion~\cite{xu2022versatile}, a model conditioned on both vision and language representations acquired from the pretrained CLIP~\cite{radford2021learning} model. 

Our framework consists of two stages. The first stage, illustrated in Figure~\ref{fig:vdvae}, generates a low-level reconstruction of images (akin to an ``initial guess'') using a Very Deep Variational Autoencoder (VDVAE)~\cite{child2021very}. We generate these reconstructions by training a regression model to associate fMRI signals to the corresponding latent variables of VDVAE for the same training images. In the second stage, illustrated in Figure~\ref{fig:vd}, we train two additional regression models: one from fMRI patterns to CLIP-Vision features (extracted by feeding the corresponding images to the CLIP model); and the other one from fMRI patterns to CLIP-Text features (collected by providing to the CLIP model the captions of the corresponding images). Finally, we use the multimodal dual-guidance as well as the image-to-image abilities of the pretrained Versatile Diffusion (VD) model to  generate the final reconstructions for test images. Using our trained regression models, for each test fMRI pattern we obtain an ``initial guess'' image (stage 1, VDVAE reconstruction) used by VD's image-to-image pipeline, as well as predicted CLIP-Vision and CLIP-Text feature vectors (stage 2), jointly used for conditioning VD's diffusion process. We used VDVAE, CLIP, and Versatile Diffusion with their pretrained weights, and did not apply any finetuning. We only trained regression models that transform fMRI patterns to latent variables of the models.

We demonstrate below that the resulting scene images reconstructed by the Brain-Diffuser model are highly naturalistic and retain the overall layout and semantic information of the groundtruth images while showing only minor variations in finer details. Compared to earlier models that exhibited proficiency in capturing certain features of groundtruth images, Brain-Diffuser demonstrates qualitatively and quantitatively superior performance in terms of both high-level and low-level metrics, thus establishing itself as state-of-the-art.

\section{Materials and Methods}

\subsection{Dataset}

We used the publicly available Natural Scenes Dataset (NSD), a large-scale 7T fMRI dataset~\cite{allen2022massive}. The NSD was collected from 8 subjects viewing images from the COCO~\cite{lin2014microsoft} dataset. Each image was viewed for 3 seconds, while subjects were engaged in a continuous recognition task (reporting whether they had seen the image at any previous point in the experiment). For our study, we used the 4 subjects (sub1, sub2, sub5, sub7) who completed all trials. The training set that we used thus contained 8859 images and 24980 fMRI trials (up to 3 repetitions for each image), and 982 images and 2770 fMRI trials for the test set. We averaged fMRI trials for the images that had multiple repetitions. We also used the corresponding captions from the COCO dataset. Test images are common for all subjects, while training images are different. We used the provided single-trial beta weights, obtained using generalized linear models with fitted hemodynamic response functions and additional GLMDenoise and ridge regression procedures (‘betas\_fithrf\_GLMdenoise\_RR’). We masked preprocessed fMRI signals using the provided NSDGeneral ROI (Region-of-Interest) mask in 1.8 mm resolution. The ROI consists of [15724, 14278, 13039, 12682] voxels for the 4 subjects respectively, and includes many visual areas from the early visual cortex to higher visual areas. For further details on this dataset and the corresponding fMRI preprocessing steps, we refer the reader to the initial paper describing the Natural Scenes Dataset~\cite{allen2022massive}.

\subsection{Low-Level Reconstruction of Images using VDVAE (first stage)}
\begin{figure}[!htb]
    \centering
    \includegraphics[width=0.6\textwidth]{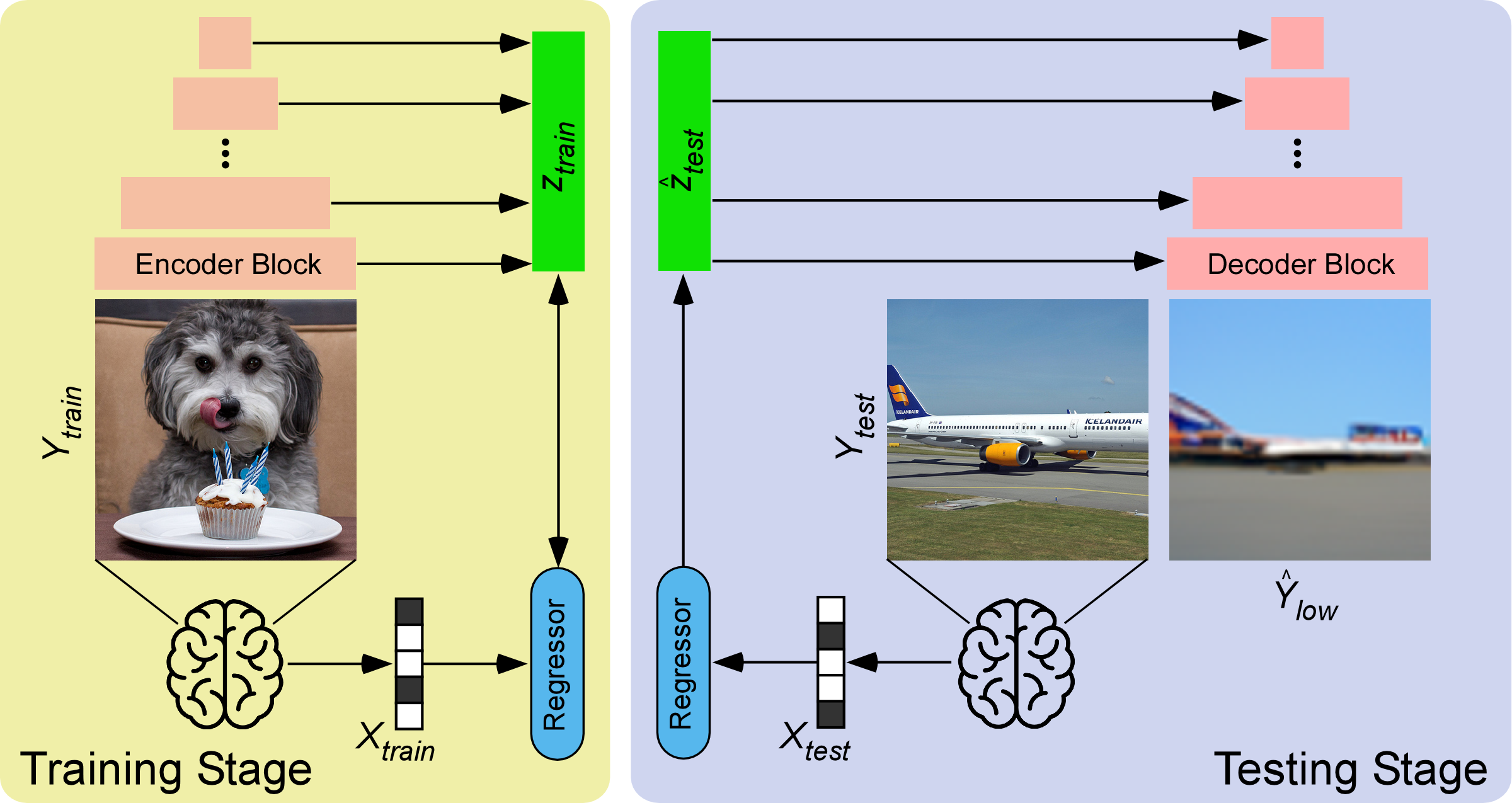}
    \caption{Reconstruction of Images via VDVAE (first stage). Training Stage (left). Latent variables ($z_{train}$) are extracted and concatenated for the first 31 layers of the hierarchy by passing training images ($Y_{train}$) into the pretrained VDVAE Encoder. A ridge regression model (Regressor) is trained between fMRI patterns ($X_{train}$) and corresponding latent variables ($z_{train}$). Testing Stage (right). Test fMRI data ($X_{test}$) are passed through the trained Regressor to obtain predicted latent variables ($\hat{z}_{test}$). These predicted latent variables are fed to the pretrained VDVAE Decoder to get the low-level reconstruction ($\hat{Y}_{low}$) of the test images ($Y_{test}$), which will serve as a sort of ``initial guess'' for the second stage. Note that all VDVAE layers (encoder and decoder blocks) are pretrained and frozen, only the brain-to-latent regression layer (blue box) is trained.}
    \label{fig:vdvae}
\end{figure}

A Variational Auto-Encoder (VAE)~\cite{kingma2013auto} is a generative model trained to capture an input distribution (such as an image dataset) via a low-dimensional latent space, constrained to follow a particular prior distribution (e.g. Gaussian). When the input dataset takes on a more complex distribution, training a Variational Autoencoder (VAE) can be challenging. Indeed, prior work has found that datasets consisting of natural scene images require many latent variables with complex distributions for which a simple VAE would not suffice; this is why the Very Deep Variational Autoencoder (VDVAE) was introduced~\cite{child2021very}. The VDVAE is a hierarchical VAE model, with several layers of conditionally dependent latent variables, each layer adding different details from coarse to fine when transitioning from top to bottom. The hierarchical dependence can be seen in equations (1) and (2), where $z$ indicates latent representations, $x$ is the input variable, $q_\phi$ represents the approximate posterior distribution that is learned when training the encoder, and $p_\theta$ represents the prior distribution that is learned when training the decoder. The latent variable $z_0$ is at the top of the hierarchy with the smallest dimension (low resolution, with coarse details) and $z_N$ is at the bottom of the hierarchy with the largest dimension (high resolution, with fine details). Equation (1) shows that the latent variables at the bottom of the hierarchy are dependent on those who are at the top (and on the input $x$). When there is no input ($x$), it is still possible to generate samples using the prior distribution described in equation (2). This hierarchical structure helps the VDVAE learn sufficiently expressive latent variables to represent complex distributions like natural scene images.

\begin{equation}
q_\phi(\boldsymbol{z} \mid \boldsymbol{x})=q_\phi\left(\boldsymbol{z}_0 \mid \boldsymbol{x}\right) q_\phi\left(\boldsymbol{z}_1 \mid \boldsymbol{z}_0, \boldsymbol{x}\right) \ldots q_\phi\left(\boldsymbol{z}_N \mid \boldsymbol{z}_{<N}, \boldsymbol{x}\right)
\end{equation}

\begin{equation}
p_\theta(\boldsymbol{z})=p_\theta\left(\boldsymbol{z}_0\right) p_\theta\left(\boldsymbol{z}_1 \mid \boldsymbol{z}_0\right) \ldots p_\theta\left(\boldsymbol{z}_N \mid \boldsymbol{z}_{<N}\right)
\end{equation}

For our study, we used the model provided in~\cite{child2021very}, trained on a $64\times64$ resolution ImageNet dataset, and consisting of 75 layers; we only utilized the latent variables from the first 31 layers for the sake of size in regression, since we observed that adding further layers did not make much difference in the reconstruction results (at test time, the latent variables from the remaining layers are sampled according to the prior distribution given in equation (2)). 

In the training stage, we fed images to the encoder part of the VDVAE to extract latent variables for each training image (as described in Figure~\ref{fig:vdvae}). We concatenated the latent variables from the 31 layers, which resulted in $91168$-dim vectors. Then, we trained a ridge regression model between fMRI training patterns and these concatenated vectors. In the testing stage, we provided test fMRI patterns to the trained regression model and thus predicted latent values for each test image. Then, we fed those latent values to the decoder part of the VDVAE and obtained reconstructed images ($64\times64$ pixels) from the VDVAE. These low-level reconstructions served as an ``initial guess'' for the diffusion model (second stage).

\subsection{Final Reconstruction of Images using Versatile Diffusion (second stage)}

\begin{figure}[!htb]
    \centering
    \includegraphics[width=0.6\textwidth]{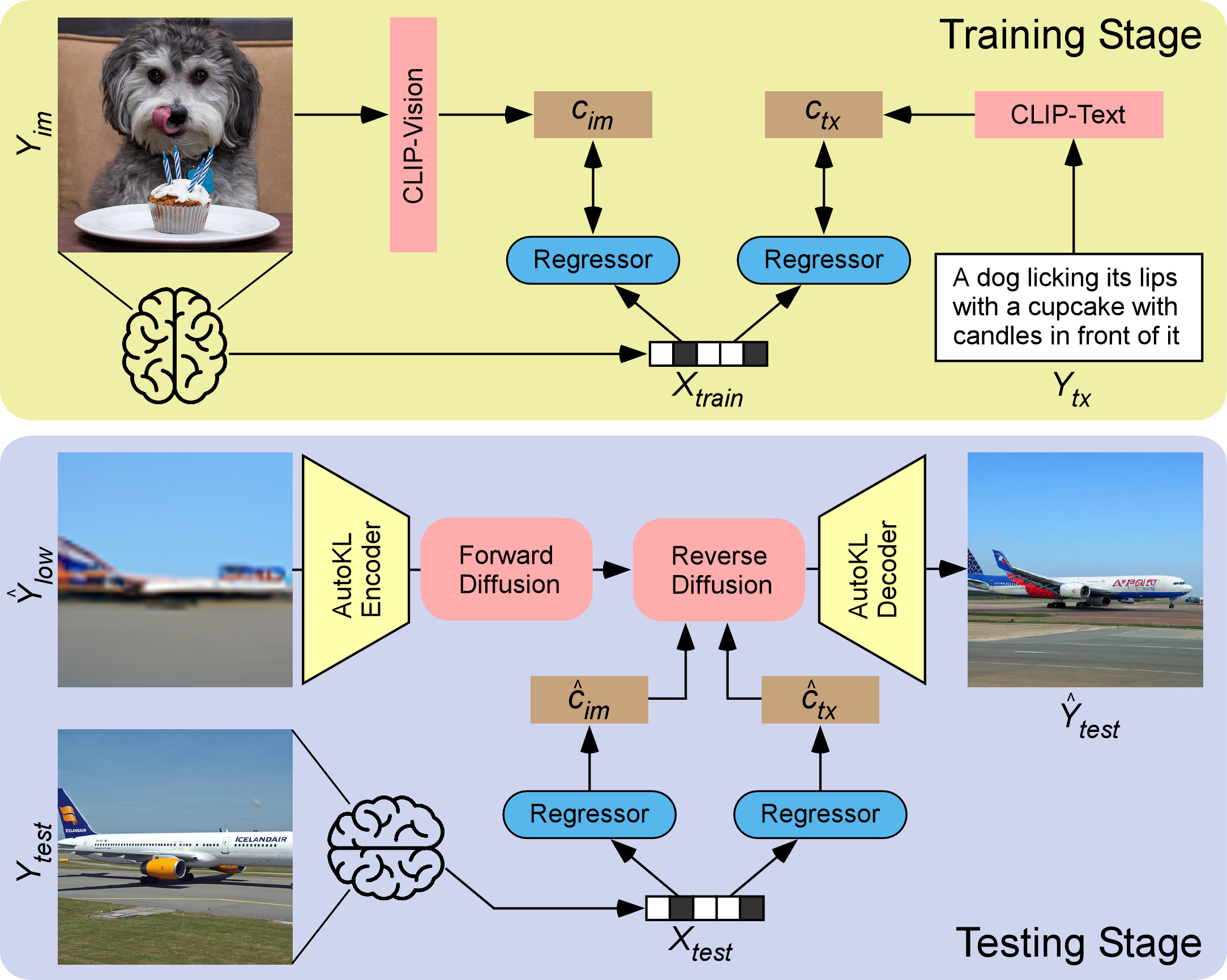}
    \caption{Final Reconstruction of Images via Versatile Diffusion (second stage). Training Stage: CLIP-Vision features ($c_{im}$) are extracted by feeding training images ($Y_{im}$) to the pretrained CLIP model. CLIP-Text features ($c_{tx}$) are extracted by providing the corresponding captions ($Y_{tx}$) to the pretrained CLIP Model. Two different ridge regression models (Regressors) are trained to learn the mapping between these features and fMRI patterns ($X_{train}$). Testing Stage: Predicted CLIP-Vision ($\hat{c}_{im}$) and CLIP-Text ($\hat{c}_{tx}$) features are computed by giving test fMRI patterns ($X_{test}$) to the trained regression models. In the image-to-image pipeline of the latent diffusion model, VDVAE reconstructions of test images (the ``initial guess'' $\hat{Y}_{low}$ from the first stage) are passed through the AutoKL Encoder of the pretrained Versatile Diffusion model, and the obtained latent vectors undergo 37 steps of the forward diffusion process (noise addition). The resulting noisy latent vectors are used to initialize the reverse diffusion process, which is also guided by predicted CLIP-Vision ($\hat{c}_{im}$) and CLIP-Text ($\hat{c}_{tx}$) features jointly in a dual-guided framework. At last, the resulting denoised latent vector is passed through the AutoKL Decoder to generate the final reconstructed image ($\hat{Y}_{test}$). Note that all CLIP (vision and text encoders) and Versatile Diffusion layers (AutoKL encoder and decoder, forward and reverse diffusion blocks) are pretrained and frozen, only the brain-to-latent regression layers (blue boxes) are trained. }
    \label{fig:vd}
\end{figure}

Although the VDVAE was helpful to reconstruct the layout of the image, it is not sufficient for the high-level features, nor does it produce fully naturalistic pictures. For that, we use the Versatile Diffusion~\cite{xu2022versatile} model in the second stage of our reconstruction framework. Versatile Diffusion is a recently proposed latent diffusion model (LDM)~\cite{rombach2022high}. 

LDMs have become highly popular after their success in high-resolution text-to-image generation. In order to train an LDM, first an autoencoder (with encoder $E(.)$ and decoder $D(.)$) is trained on a large-scale image dataset to learn a compressed representation of images $x_0$, i.e. a latent space $z_0=E(x_0)$. Then, the forward diffusion process is applied to these latent variables $z_0$ by adding Gaussian noise in successive timesteps (described in equation (3), where $t$ represents the timestep, $\bar{\alpha}_t$ indicates a coefficient derived from the standard deviation of the Gaussian noise, and $\varepsilon$ represents the Gaussian noise). The reverse diffusion process is learned via a neural network (Denoising U-Net in the original paper) to predict and remove noise from the noisy latent so as to retrieve the original latent variables. This is done by minimizing the loss function in equation (4), where $\varepsilon$ is the true Gaussian noise, $\varepsilon_\theta(.)$ represents the neural network being trained to predict the noise, $z_t$ is the latent variable, $t$ is the timestep, and $\tau_\theta(y)$ is the conditioning input for the U-Net. After the reverse diffusion process, the denoised latent variables are passed through the trained decoder $D(.)$ to generate the images. The critical part of this process is that it is possible to condition this reverse diffusion process on different representations (e.g text captions, images, semantic maps). This conditioning process is done by merging conditions ($\tau_\theta(y)$) in the cross-attention block of the Denoising U-Net. 
\begin{equation}
z_t=\sqrt{\bar{\alpha}_t} z_0+\sqrt{1-\bar{\alpha}_t} \varepsilon
\end{equation} 
\begin{equation}
L_{\mathrm{LDM}}=\mathbb{E}_{t, z_0, \varepsilon, y}\left[\left\|\varepsilon-\varepsilon_\theta\left(z_t, t, \tau_\theta(y)\right)\right\|^2\right]
\end{equation}

The Versatile Diffusion model (see Figure~\ref{fig:vd}) is a latent diffusion model with different pathways which allow us to condition the generation process on both text and image features to guide the reverse diffusion process. It is possible to provide CLIP-Vision, CLIP-Text, or both features as conditions in the reverse diffusion process. It is also possible to initialize the reverse diffusion with latent variables obtained from a particular image, rather than from a purely random distribution--this is the image-to-image pipeline that we will use to take advantage of our ``initial guess'' image from stage 1.  The Versatile Diffusion model that we utilized in our framework was trained on the Laion2B-en~\cite{schuhmann2021laion} dataset with $512\times512$ resolution images and corresponding captions. CLIP (Contrastive Language-Image Pre-training)~\cite{radford2021learning} is a multimodal model designed to assist in different tasks that involve natural language processing and computer vision. It is trained in a contrastive learning approach, where features gathered from images vs. text captions are projected onto separate latent spaces of identical dimensions: CLIP-V refers to the latent space for images and CLIP-T for captions. Similarity scores (e.g. cosine similarity) of the latent space projections for matching images and captions are optimized throughout training. CLIP is widely used as a feature extractor, due to its high representational capabilities. The CLIP network used in Versatile Diffusion is based on the transformer architecture (ViT-L/14) and pretrained on a large-scale contrastive task.

In stage 2, we thus train two regression models, the first one between fMRI patterns and CLIP-Vision features (with $257\times768$-dim extracted from the corresponding images where the first vector with $768$-dim represents the category-related embedding and the remaining $256$ embeddings represent the patches acquired from the images) and the second one between fMRI patterns and CLIP-Text features ($77\times768$-dim extracted from the COCO captions associated with the corresponding images where the $77$ embeddings correspond to the number of tokens given to the model as inputs). At testing time, we use the image-to-image pipeline of the latent diffusion model. First, we encode the image reconstructed with the VDVAE model (stage 1) with the AutoKL Encoder (after upsampling the image from $64\times64$ to $512\times512$) and add noise to the latent vector for 37 steps of forward diffusion (corresponding to 75\% of the 50 steps of full diffusion, which is a commonly used value in the image-to-image pipeline of LDMs.). In this image-to-image pipeline, it is necessary to first add some amount of noise to the latent values using forward diffusion, since LDMs generate images via denoising using reverse diffusion (without noise on the image, the reverse diffusion step would end up with no change). Then, we feed this noisy latent as initialization to the diffusion model and denoise it for 37 steps while conditioning with the predicted CLIP-Vision and CLIP-Text features (stage 2). In every step of reverse diffusion, we use CLIP-Vision and CLIP-Text jointly in the double-guided diffusion pipeline of Versatile Diffusion, where the cross-attention matrices for both conditions are mixed through linear interpolation (with CLIP-Vision having a relative strength of 0.6 and CLIP-Text of 0.4). The diffusion result is passed through the AutoKL Decoder to produce our final $512\times512$ pixel reconstruction. 

\subsection{Code availability}
The code for our project, including scripts to train regression models, pretrained weights, and scripts to produce reconstructions for test images and for ROI-based synthetic patterns, is publicly available at github.com/ozcelikfu/brain-diffuser.

\section{Results and Analyses}
\subsection{Image reconstruction examples}

\begin{figure*}[!htb]
    \centering
    \includegraphics[width=0.49\textwidth]{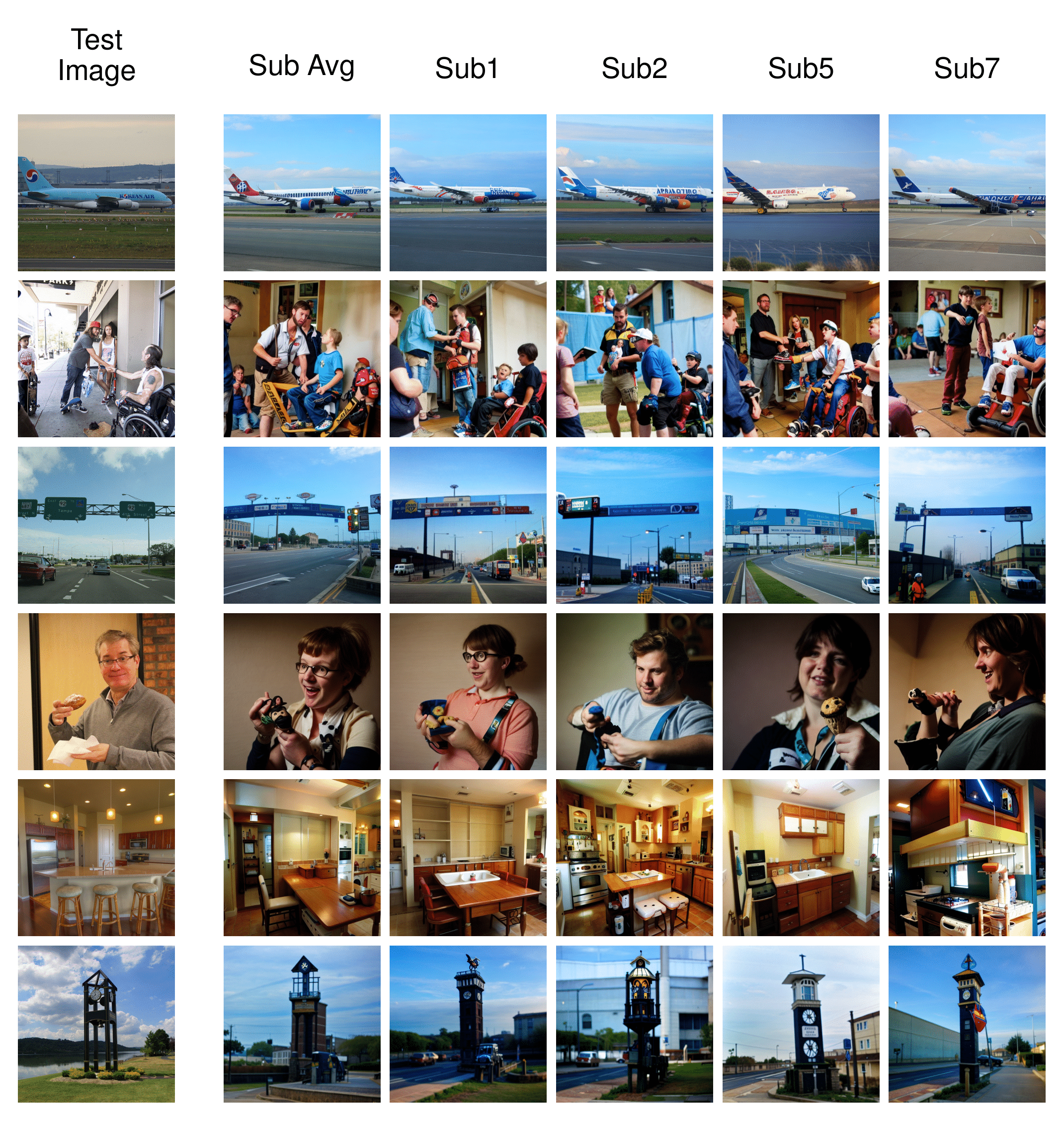}
    \includegraphics[width=0.49\textwidth]{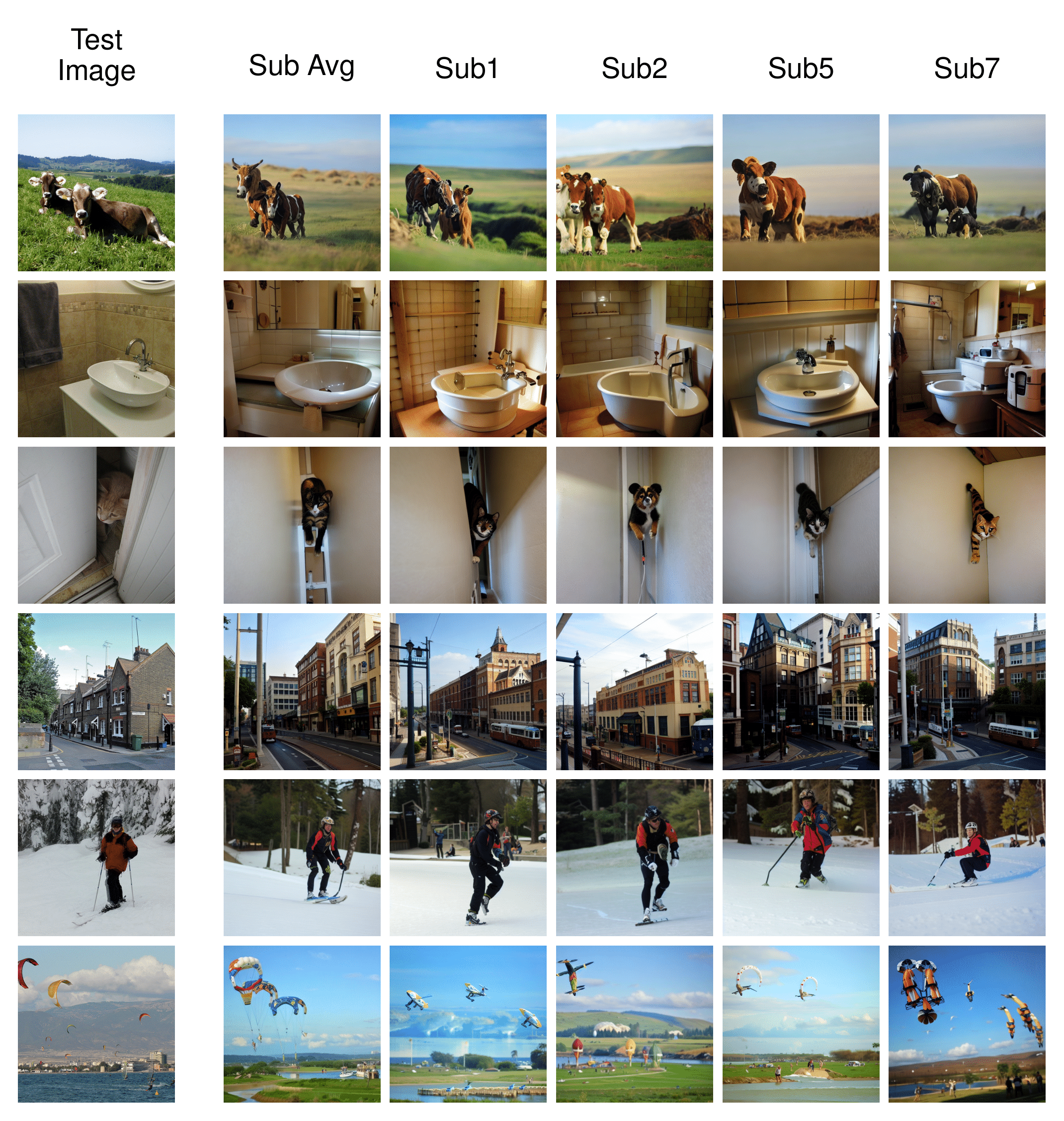}
    \caption{Examples of fMRI Reconstructions from our Brain-Diffuser model. The first column is the groundtruth image (Test Image). The second column is generated by averaging the predicted latent variables over all 4 subjects seeing the same picture (Sub Avg). The remaining columns are for each individual subject (Sub1, Sub2, Sub5, Sub7)}
    \label{fig:multi_subject}
\end{figure*}

We present examples of reconstructions from our model in Figure~\ref{fig:multi_subject}. While we present the results of each individual subject in different columns, we also added results gathered by averaging the latent variables predicted by all subjects. In general, we see that reconstructed images capture most of the layout and semantics of the groundtruth images, while there remain differences in pixel-level details. For instance, looking specifically at the first four images on the left, we see that the reconstructed pose (3D orientation) of the plane (first image) is correct for every subject although there are some differences in the details of the plane and also in the texture of the background. Nonetheless, the fact that a commercial plane on a runway, facing to the right on a blue sky background was reconstructed in all instances is not a trivial feat.
For the second example, all reconstructed images display a group of people, although layouts tend to differ. Still, a person in a wheelchair is visible in the bottom right corner for three of the four subjects.
For the third image, the model reconstructed a highway with road signs correctly, although the orientation of the road is different for some of the subjects, and the details of the signs are not entirely matched.
On the fourth sample, all reconstructed images show a single person facing left and holding an object in their hand, as in the groundtruth image. The person's details (gender, age, clothing) are different across subjects, e.g. with glasses only reconstructed for subject 1 and in the average across subjects. Reconstructed image contrast also differs from the ground truth. 
Similar conclusions can be generalized to most images of the test set: while never passing for a picture-perfect copy, with visible differences in especially color and contrast (due to inherent limitations of the Versatile Diffusion model in this respect), the reconstructed images are always naturalistic (that is, as much as diffusion models can generate) and plausible alternate renditions of the ground truth.

\begin{figure*}[!htb]
    \centering
    \includegraphics[width=0.40\textwidth]{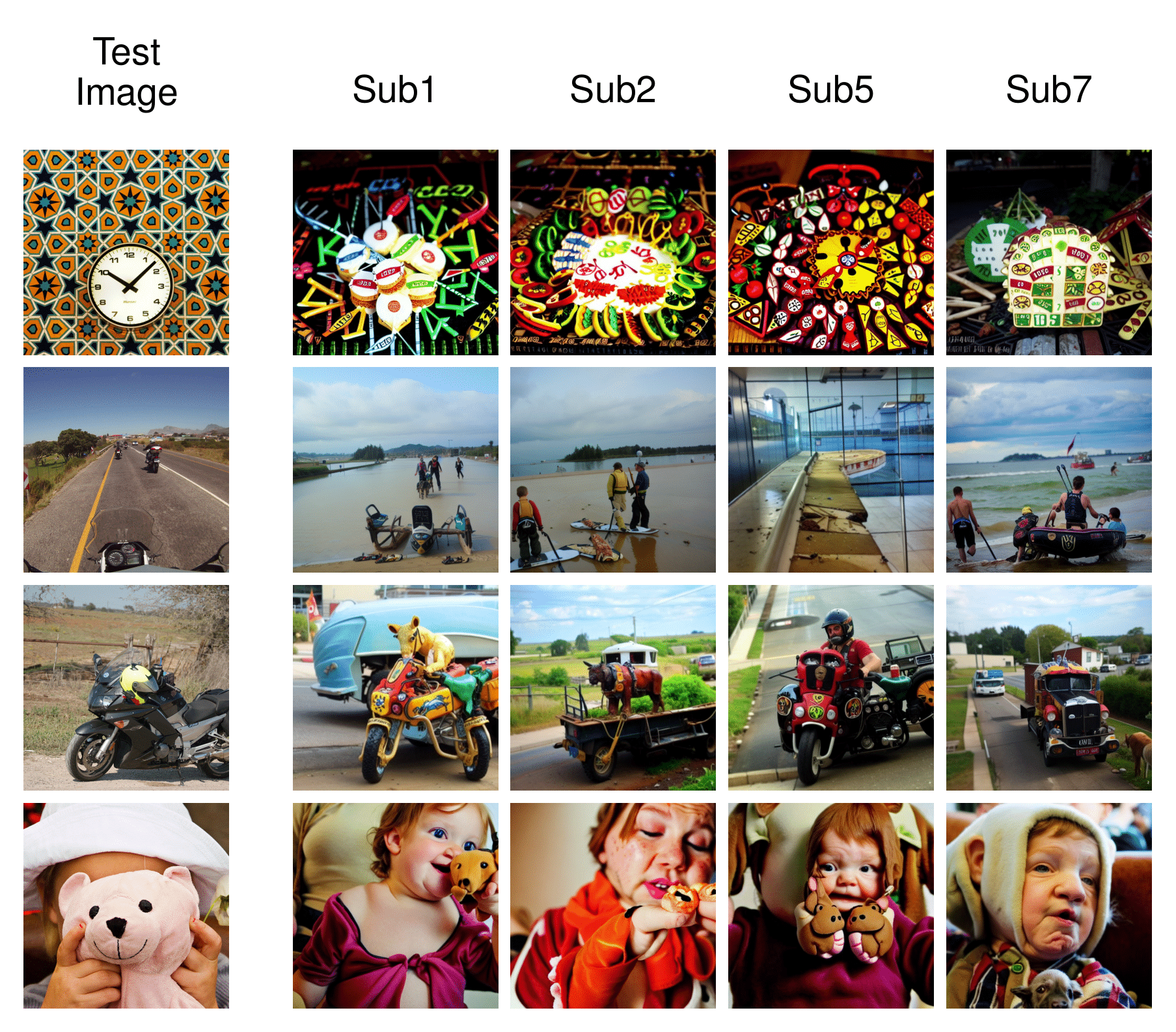}
    \includegraphics[width=0.40\textwidth]{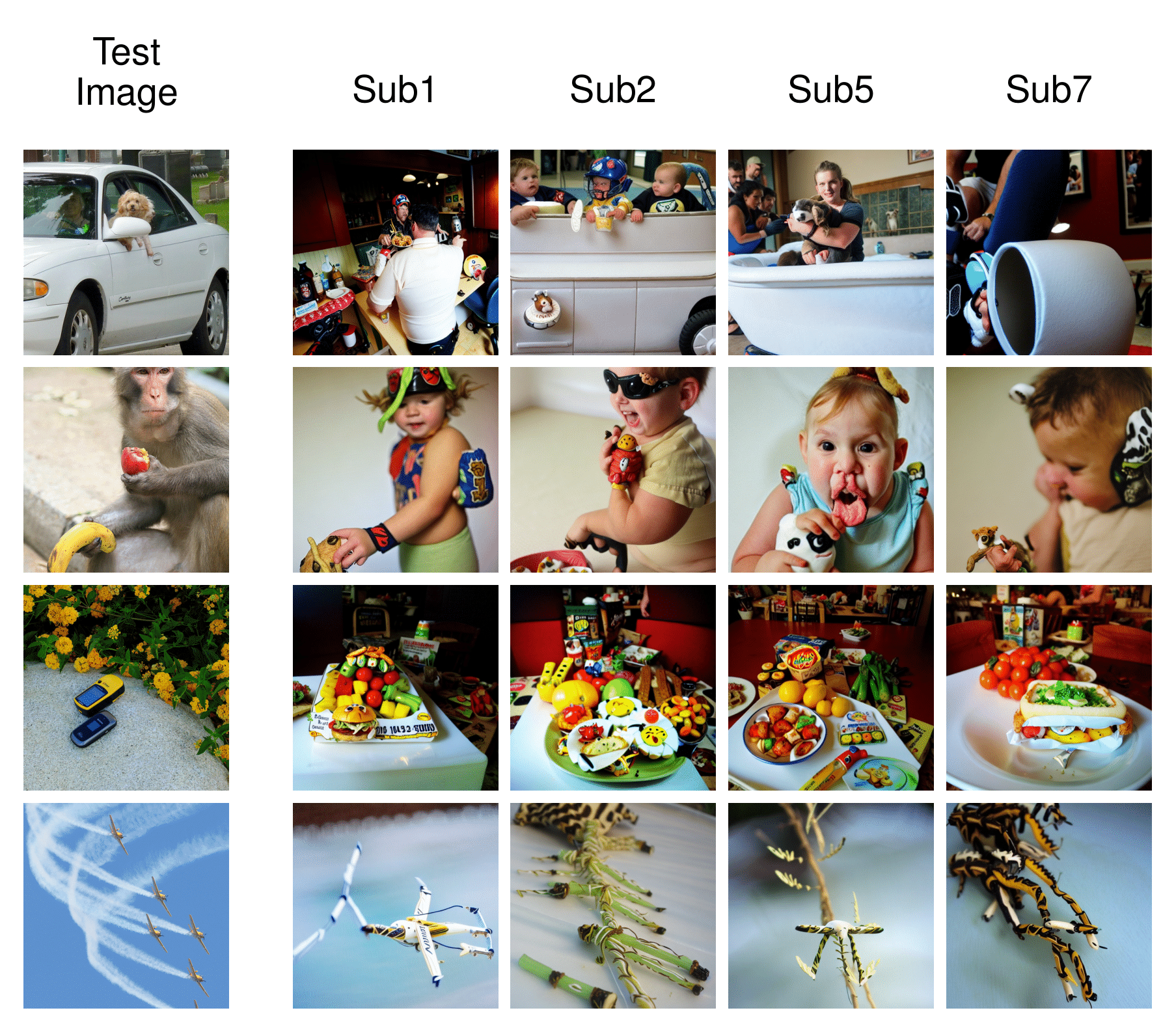}
    \caption{Failure cases of fMRI Reconstructions from our Brain-Diffuser model. The first column is the groundtruth image (Test Image). The remaining columns are for each individual subject (Sub1, Sub2, Sub5, Sub7)}
    \label{fig:failure_case}
\end{figure*}

We also present some examples of reconstruction failures from our model in Figure~\ref{fig:failure_case}. In these examples, we see that our model can fail due to different reasons. In the first example, although Brain-diffuser reconstructs oval objects around the center, the complex texture of the background seems to interfere with the object, which is not reconstructed as a clock. For the second example, the reconstructions show sea in the background, although there is no sea in the ground-truth image. On the fourth sample, the teddy bear occluding the kid's face seems to confuse the model, as it generates human faces in the reconstructions. For the sixth example, Brain-Diffuser reconstructs a kid instead of a monkey. These examples highlight the fact that Brain-Diffuser can fail on occasion, due to diverse reasons like complex stimuli, object occlusions, or confusing one object with another.

\begin{figure*}[!htb]
    \centering
    \includegraphics[width=0.75\textwidth]{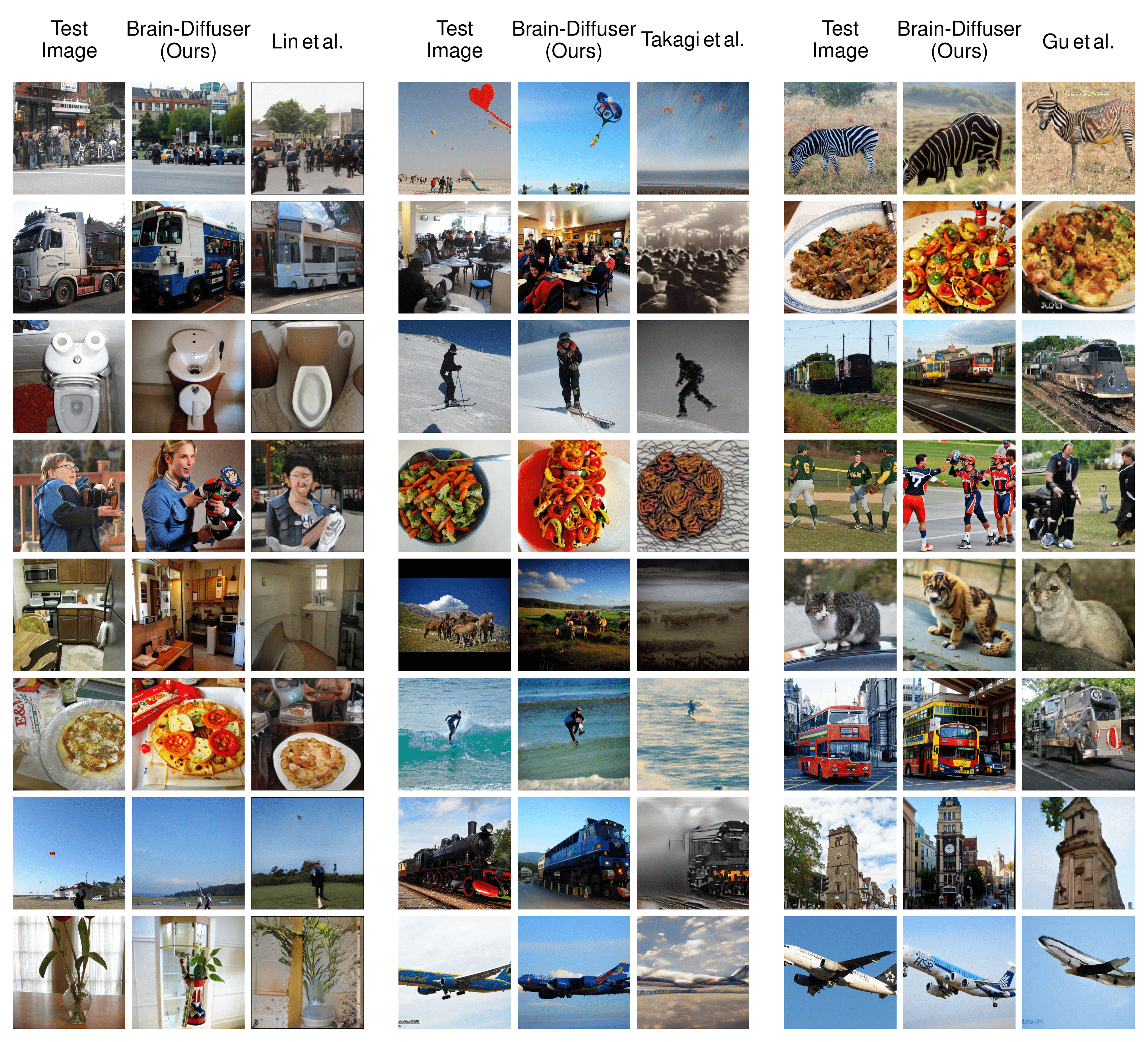}
    \caption{Comparison of fMRI Reconstructions for different models. Since the presented test images in all methods were different, we did comparisons separately for each model. On the left (first 3 columns), we present the comparison of our model with Lin et al. together with groundtruth test images. On the center (columns 4-6), we present the comparison of our model with Takagi et al. together with groundtruth test images. On the right (last 3 columns), we present the comparison of our model with Gu et al. together with groundtruth test images.   }
    \label{fig:method_comparison}
\end{figure*}

\subsection{Comparison with state of the art}
How do these findings compare to the state of the art? We contrast the qualitative results of our model with three other existing models in Figure~\ref{fig:method_comparison}. We selected 8 images for each model comparison; we use different images each time since these studies presented different images in their papers. (Note that for the specific purpose of comparing with Lin et al., we re-trained and re-tested our model with a different training and testing split, to match the split that they had used). 

Lin et al.~\cite{lin2022mind} was the first study that used the NSD dataset for reconstruction. They are similar to our model in terms of utilizing both image and text features as conditions, but they used a StyleGAN2 model instead of an LDM. Although Lin et al seem to be performing better than the other two prior models, in some instances the quality of their reconstructions still lags behind ours. For instance, in the second image, the details of the truck are better represented in our model, while for the third image, the shape of the toilet is better represented in Lin et al. In the fourth image, the color of the clothes is presented more accurately in our model, as well as the fact that the person is holding an item; the person's face also looks more realistic compared to Lin et al. On the other hand, the color and location of the pizza in the sixth image appear more aligned with the ground-truth image for Lin et al.
Takagi et al.~\cite{takagi2023high} is the only other study (in addition to ours) to use a latent diffusion model for reconstructing images from the NSD dataset. Although their model generates easily recognizable silhouettes, they do not seem to perform as well as our model in any qualitative aspect including low-level details, semantics, or naturalness. 
Finally, Gu et al.~\cite{gu2023decoding} used an Instance-Conditioned GAN model trained on ImageNet. When we compare our results to theirs, we can see that, although both appear good at reconstructing images with similar semantics, structural aspects are less well represented in their reconstructed images (e.g. unrealistic warped shapes for the train, bus, and building). In contrast, the shape and texture details of our model are more realistic. Since their model has a BigGAN backbone, with few parameters to encode the entire layout of the image (including the object's class, its pose, size, and location), and since it is trained on a single-object-centric dataset (ImageNet), the model seems to be limited in reconstructing complex scenes with multiple objects. On the other hand, since LDMs include a spatially organized map of features, it is more convenient for them to represent multiple objects; as an example, we see one train in the third image although there are two trains in the groundtruth image, and in the reconstructed image from our model.

\begin{table*}[htbp]
    \centering
    \caption{Quantitative Analysis of fMRI Reconstructions. For each measure, the best value is in bold. (For PixCorr, SSIM, AlexNet(2), AlexNet(5), Inception and CLIP metrics, higher is better. For EffNet-B and SwAV distances, lower is better. This is indicated by the arrow pointing up or down, respectively)}
    \label{tab:method_cmp}
    \small
    \begin{tabular}{|l|c|c|c|c|c|c|c|c|}
        \hline
        \multirow{3}{*}{Method} & \multicolumn{8}{c|}{Quantitative Measures}\\
        \cline{2-9}
        ~  & \multicolumn{4}{c|}{Low-Level} & \multicolumn{4}{c|}{High-Level}\\
        \cline{2-9}
        ~ & PixCorr $\uparrow$ & SSIM $\uparrow$ & AlexNet(2) $\uparrow$ & AlexNet(5) $\uparrow$ & Inception $\uparrow$ & CLIP $\uparrow$  & EffNet-B $\downarrow$ & SwAV $\downarrow$\\
        \hline
        \hline
         Lin et al.~\cite{lin2022mind} & $-$ & $-$ & $-$ & $-$  & $78.2\%$ & $-$ & $-$ & $-$\\
         Takagi et al.~\cite{takagi2023high} & $-$ & $-$ &  $83.0\%$ & $83.0\%$  & $76.0\%$ & $77.0\%$ & $-$ & $-$\\
         Gu et al.~\cite{gu2023decoding} & $.150$& $.325$ & $-$ & $-$  & $-$ & $-$ & $.862$ & $.465$\\
         Brain-Diffuser (Ours)  & $\mathbf{.254}$ & $\mathbf{.356}$ & $\mathbf{94.2\%}$ & $\mathbf{96.2\%}$  & $\mathbf{87.2\%}$ & $\mathbf{91.5\%}$ & $\mathbf{.775}$ & $\mathbf{.423}$\\
         \hline
        
    \end{tabular}
    
\end{table*}

\subsection{Quantitative Results}

To make the comparison with other models more quantitative, we present the results of 8 different image quality metrics in Table~\ref{tab:method_cmp}. PixCorr is the pixel-level correlation of reconstructed and groundtruth images. SSIM\cite{wang2004image} is the structural similarity index metric. AlexNet(2) and AlexNet(5) are the 2-way comparisons of the second and fifth layers of AlexNet~\cite{krizhevsky2017imagenet}, respectively. Inception is the 2-way comparison of the last pooling layer of InceptionV3~\cite{szegedy2016rethinking}. CLIP is the 2-way comparison of the output layer of the CLIP-Vision~\cite{radford2021learning} model. EffNet-B and SwAV are distance metrics gathered from EfficientNet-B1~\cite{tan2019efficientnet} and SwAV-ResNet50~\cite{caron2020unsupervised} models, respectively. The first four can be considered as low-level metrics, while the last four reflect higher-level properties. For PixCorr and SSIM metrics, we downsampled generated images from $512\times512$ resolution to $425\times425$ resolution (i.e. the resolution of groundtruth images in NSD dataset). For the rest of the measures, generated images are preprocessed according to the input properties of each network. Note that not all measures are available for each previous model (depending on what they chose to report). However, each model has at least one point of comparison with ours. Our quantitative comparisons with Takagi et al. and Gu et al. are made according to the exact same test set, i.e., the 982 images that are common for all 4 subjects. Lin et al., on the other hand, reported their results on only Subject 1 and with a custom train-test set split. However, when measuring our model's image quality on the same train-test split as Lin et al, we observed nearly identical results (Inception Score of 87.0\%, compared to 78.2\% for Lin et al). Our model is the best-performing model by a decent margin for all of the quantitative metrics. Overall, these results show that our model can be considered state-of-the-art for both low-level and high-level quantitative measures. 

\subsection{Ablation Studies}

\begin{table*}[htbp]
    \centering
    \caption{Quantitative comparisons of test fMRI reconstructions of Sub1 with various ablations of the full model. For each measure, the best value is in bold. (For PixCorr, SSIM, AlexNet(2), AlexNet(5), Inception, and CLIP metrics, higher is better. For EffNet-B and SwAV distances, lower is better. This is indicated by the arrow pointing up or down, respectively)}
    \label{tab:ablation_cmp}
    \small
    \begin{tabular}{|l|c|c|c|c|c|c|c|c|}
        \hline
        \multirow{3}{*}{Method} & \multicolumn{8}{c|}{Quantitative Measures}\\
        \cline{2-9}
        ~  & \multicolumn{4}{c|}{Low-Level} & \multicolumn{4}{c|}{High-Level}\\
        \cline{2-9}
        ~ & PixCorr $\uparrow$ & SSIM $\uparrow$ & AlexNet(2) $\uparrow$ & AlexNet(5) $\uparrow$ & Inception $\uparrow$ & CLIP $\uparrow$  & EffNet-B $\downarrow$ & SwAV $\downarrow$\\
        \hline
        \hline
         Only-VDVAE & $\mathbf{.358}$ & $\mathbf{.437}$ & $\mathbf{97.7\%}$ & $\mathbf{97.6\%}$  & $77.0\%$ & $71.1\%$ & $.906$ & $.581$\\
         Brain-Diffuser w/o VDVAE & $.143$ & $.302$ & $85.6\%$ & $93.0\%$  & $87.3\%$ & $\mathbf{92.6\%}$ & $.775$ & $\mathbf{.414}$\\
         Brain-Diffuser w/o CLIP-Text & $.279$ & $.333$ &$95.6\%$ & $97.0\%$  & $\mathbf{87.9\%}$ & $91.2\%$ & $.796$ & $.436$\\
         Brain-Diffuser w/o CLIP-Vision & $.327$ & $.433$ & $93.9\%$ & $94.1\%$  & $84.7\%$ & $84.5\%$ & $.821$ & $.509$\\
         Brain-Diffuser & $.305$ & $.367$ & $96.7\%$ & $97.4\%$  & $87.8\%$ & $92.5\%$ & $\mathbf{.768}$ & $.415$\\

         \hline
        
    \end{tabular}
    
\end{table*}

\begin{figure*}[!htb]
    \centering
    \includegraphics[width=0.49\textwidth]{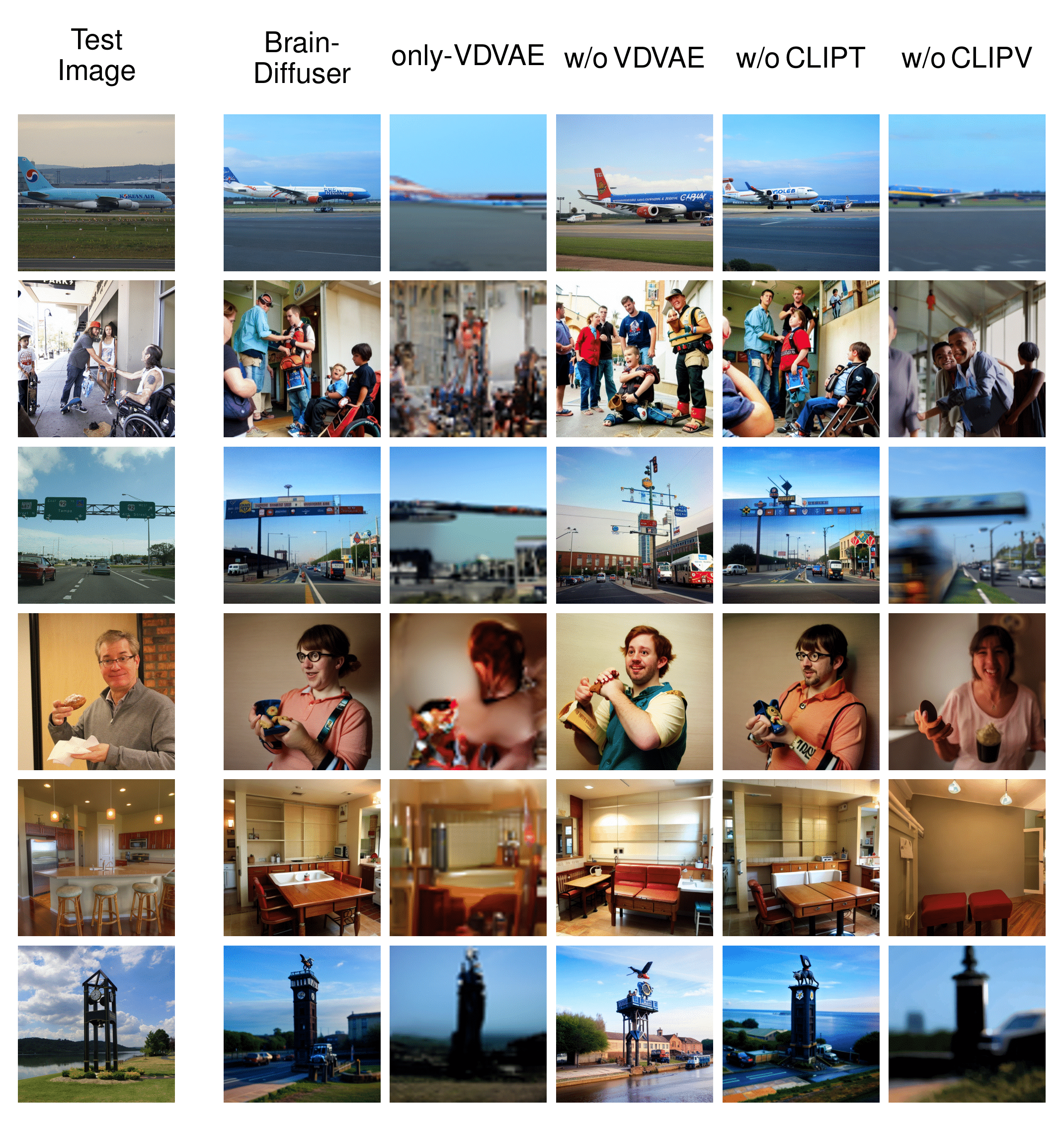}
    \includegraphics[width=0.49\textwidth]{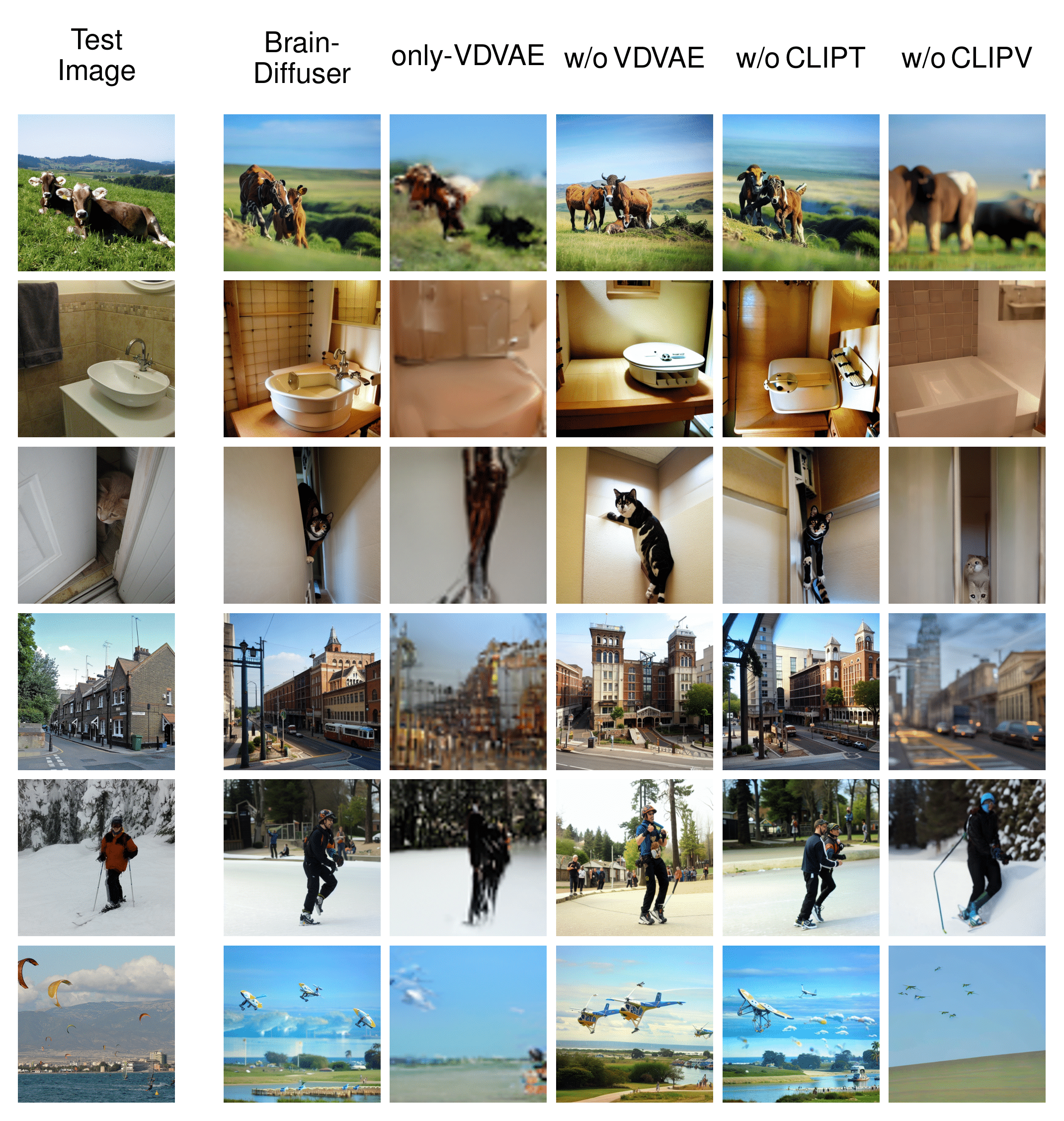}
    \caption{Examples of fMRI test reconstructions from Sub1 with various ablations of the full model. The first column is the groundtruth image (Test Image). The second column shows reconstructions from the full Brain-Diffuser model with all of its components. The third column is for reconstructions of the Only-VDVAE model. The remaining columns are for Brain-Diffuser with one of its components excluded, in order: without VDVAE, without CLIP-Text, and without CLIP-Vision.}
    \label{fig:ablation_study}
\end{figure*}

In order to reveal the contribution of each component of Brain-Diffuser, we performed an ablation study (with fMRI data of Sub1), and report both quantitative (Table~\ref{tab:ablation_cmp}) and qualitative (Figure~\ref{fig:ablation_study}) results. The quantitative results are given in Table~\ref{tab:ablation_cmp}. Our first ablation (Only-VDVAE) considers the results from stage-1 reconstruction only (Figure~\ref{fig:vdvae}) without stage-2 reconstruction (Figure~\ref{fig:vd}). This Only-VDVAE model provides the best results for all low-level measures, but the worst (by a large margin) for all high-level measures. This pattern of results is expected since the VDVAE reconstruction relies on low-level information without a contribution of semantic information from stage-2. By contrast, Brain-Diffuser without the VDVAE component (i.e., stage-2 reconstruction but with random initialization of the autoKL latent vector) performs worst on low-level measures (by a large margin), while it is among the best in high-level measures. This is also reasonable since this ablated model generates the reconstructions from only high-level features obtained from CLIP-Text and CLIP-Vision models and does not have much information about low-level information such as layout. Together, these results indicate that the VDVAE ``initial guess'' (stage-1) is necessary but not sufficient for optimal reconstruction. This is evident in the results from the full Brain-Diffuser model (last row in Table~\ref{tab:ablation_cmp}), where the contribution from VDVAE (stage 1) brings strong improvements in low-level measures, with near-optimal high-level features. In another ablation, we evaluate Brain-Diffuser without CLIP-Text. Compared to the full model, there is a sizeable decrement in both low-level and high-level measures, except Inception. While the contribution of CLIP-Text to the reconstruction of high-level semantic features is expected, its improvement of low-level measures is more surprising but could be explained by semantic information related to the image layout itself, such as the number or orientation of objects (see examples in Figure~\ref{fig:ablation_study}). Finally, Brain-Diffuser without CLIP-Vision, surprisingly, retains high performance on the low-level PixCorr and SSIM measures (lower than Only-VDVAE, but higher than the full model);  we assume that this could be due to insufficient diffusion steps (as discussed further below), preventing the reconstruction from deviating from the VDVAE initial guess. For all other measures (including low-level AlexNet measures), removing CLIP-Vision guidance severely impairs the performance of Brain-Diffuser. Overall, when jointly considering low-level and high-level measures, these quantitative results show that the full Brain-Diffuser model is better than any other variation or ablation.

We also present qualitative results in Figure~\ref{fig:ablation_study} with the same set of images presented in Figure~\ref{fig:multi_subject} of the main manuscript. Reconstructions from the Only-VDVAE model (i.e., stage-1 without stage-2) match the low-level details (e.g. shapes, layouts) of the groundtruth images, but they look like vague silhouettes rather than natural images. In contrast, Brain-Diffuser without VDVAE generates images that match high-level properties (semantics) of groundtruth images but lack positional information about the objects and their layout. This is particularly clear for the fourth image in the right part of the figure, where the layout of the street and buildings is properly captured by VDVAE (and thus, also by the full model), but is lacking in the VDVAE ablation. The images generated by Brain-Diffuser without CLIP-Text appear very close to those from the full model but with some notable differences. One example is the ski image (Row 5 on the right part of the figure), where the full model generates a single person (as in the groundtruth) while the model without CLIP-Text generates two people. Another example is the plane image (Row 1 on the left part of the figure ) where the model without CLIP-Text does not produce an image with the correctly positioned plane. Finally, reconstructions from Brain-Diffuser without CLIP-Vision appear quite blurry, and somehow in between the Only-VDVAE and the full model reconstructions. This could be an indication that forward and reverse diffusion steps were not sufficient for this model. Still, increasing the number of diffusion steps may not be a good solution since that would cause the model to lose low-level information provided by VDVAE. Overall, these qualitative examples corroborate the quantitative findings in Table~\ref{tab:ablation_cmp} and make it clear that the Brain-Diffuser model represents the optimal compromise for both low-level details and high-level semantic features.

\subsection{Which brain regions are used?}

\begin{figure}[!htb]
\centering
\includegraphics[width=0.5\textwidth]{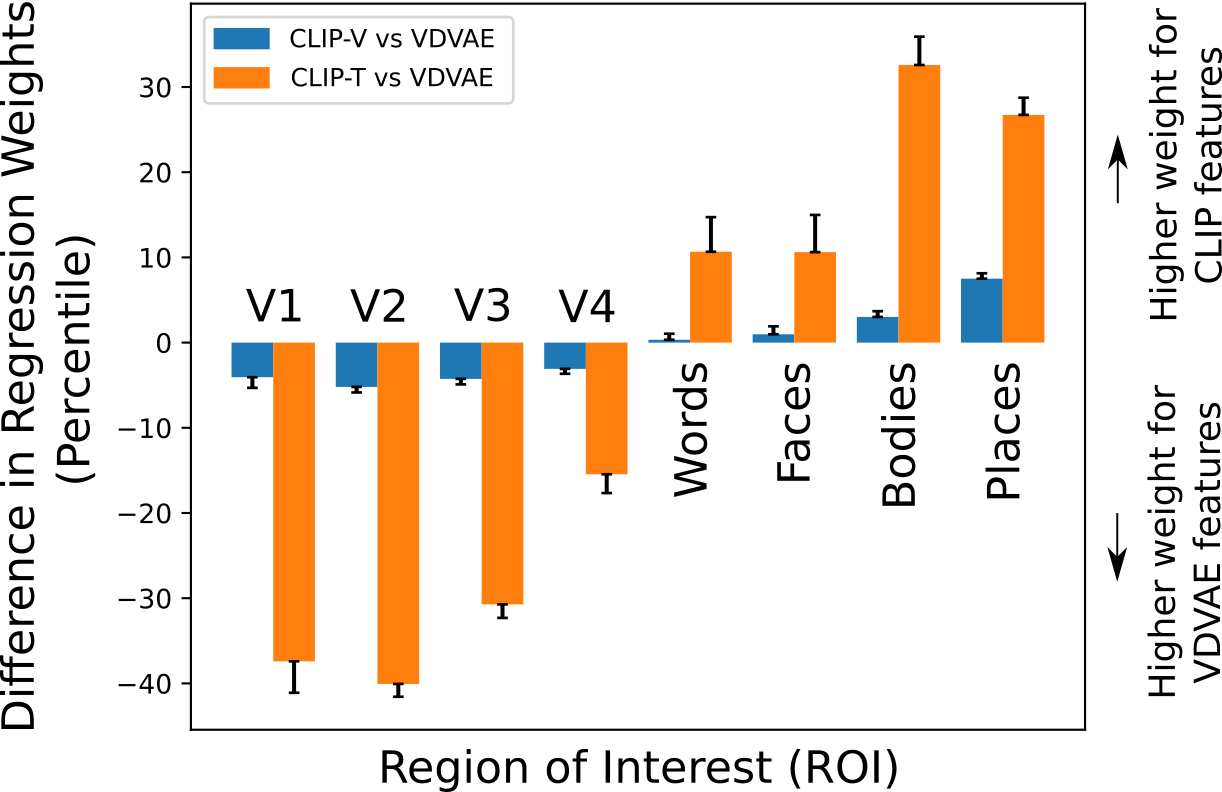}
\caption{Difference between the percentiles of the regression weights ($L_1$ norm) for the CLIP features (CLIP-V and CLIP-T) vs. the VDVAE features, averaged over voxels in each ROI and normalized by the average percentile of VDVAE features for the same ROI. Positive values indicate relatively higher regression weight for CLIP features compared to the VDVAE features, and vice versa. Error bars represent the standard error of the mean across 4 subjects.}
\label{fig:percentile_figure}
\end{figure}

In order to understand the relationship between brain regions and the various components of our model (VDVAE, CLIP-Vision, CLIP-Text), we performed a region-of-interest (ROI) analysis of the regression weights. We used 4 visual ROIs derived from population receptive field (pRF) experiments, and 4 ROIs derived from functional localization (fLoc) experiments. All experiments were provided along with the NSD dataset by the original authors. These ROIs are as follows (region names following the terminology adopted in Allen et al.~\cite{allen2022massive}): V1 is the concatenation of V1 ventral (V1v) and V1 dorsal (V1d), and similarly for V2 and V3; V4 is the human V4 (hV4); the Face-ROI consists in the union of OFA, FFA-1, FFA-2, mTL-faces, and aTL-faces; Word-ROI consists in OWFA, VWFA-1, VWFA-2, mfs-words, and mTL-words; Place-ROI consists in OPA, PPA, and RSC; Body-ROI consists in EBA, FBA-1, FBA-2, and mTL-bodies. For each voxel in these regions, we computed the strength of the regression weights ($L_1$ norm) for the CLIP features (CLIP-V and CLIP-T) and the VDVAE features, expressed as a percentile. Because the absolute regression weights can be affected by the number of voxels in each region, as well as the overall activity level and the noise level, we report our results as a \textit{difference} in regression weights between CLIP features and VDVAE features.  The results in Figure~\ref{fig:percentile_figure} show that early regions (V1-V4) are more informative about the VDVAE features, while category-specific higher brain regions (Words, Faces, Bodies, Places) carry more information about CLIP features. Another important observation is that the differences between CLIP-V and VDVAE are in the same direction, but much weaker than the differences between CLIP-T and VDVAE. This may indicate that although the Versatile Diffusion model uses CLIP-V features for high-level guidance, these features still contain more information about low-level properties than CLIP-T features.

\subsection{ROI-optimal stimuli}

\begin{figure}[!htb]
    \centering
    \includegraphics[width=0.75\textwidth]{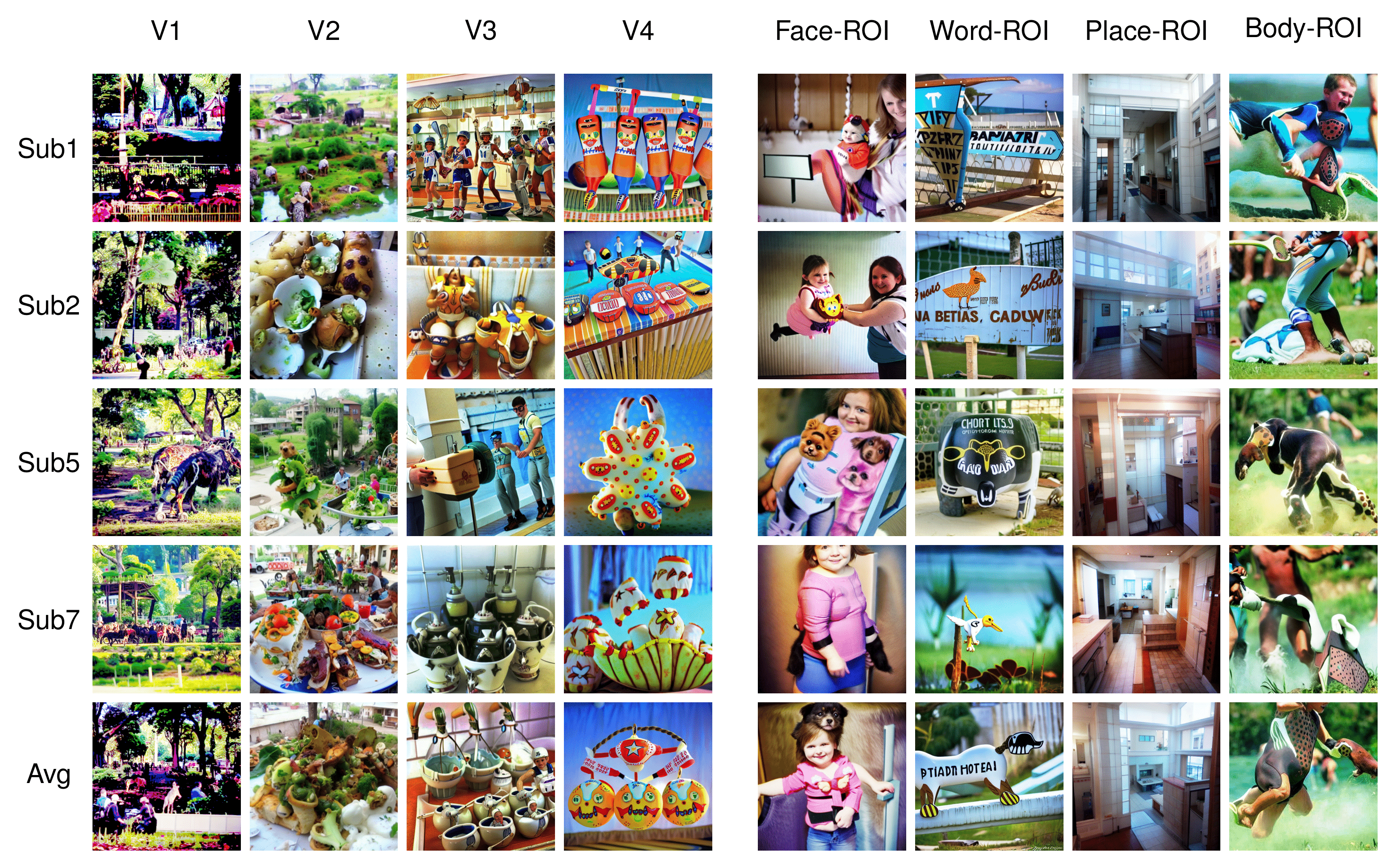}
    \caption{Images reconstructed from synthetic fMRI patterns created by activating specific regions-of-interests (ROIs). The first 4 rows present individual subjects: Sub1, Sub2, Sub5, and Sub7. The last row is generated by averaging the latent vectors predicted from all 4 subjects. The columns present ROIs: First four are ROIs from the visual cortex (V1-V4) gathered by population receptive field experiments, and the last four are ROIs that are specified with functional localization experiments (Face-ROI, Word-ROI, Place-ROI, Body-ROI). Since our synthetic fMRI patterns produce distribution shifts in the latent variables, which in turn can affect the contrast of the reconstructed images, histogram stretching and equalization are applied on color histograms of generated images for visualization purposes.}
    \label{fig:roi_analysis}
\end{figure}

Beyond brain decoding, we show here that our method can also be used to help understand the functional properties of specific regions-of-interest (ROIs) in the brain. Although we know from early studies in the neuroscience literature~\cite{hubel1962receptive,gross1972visual,perrett1982visual,gallant1993selectivity,van1994neural,kanwisher1997fusiform,epstein1998cortical,haxby2001distributed,orban2004comparative} what sort of visual properties would best activate neurons in each brain region, there are only a few studies~\cite{bashivan2019neural,ratan2021computational,gu2022neurogen,ozcelik2022reconstruction,mueller2023macaques} which attempted to directly visualize an ``optimal'' stimulus for a given brain region. Our method can easily be adapted for this purpose. We define "ROI-optimal" as images that would activate a certain ROI maximally while not activating other ROIs (or just activating them minimally). We analyzed the same 8 ROIs (V1, V2, V3, V4, Face-ROI, Word-ROI, Place-ROI, and Body-ROI) that we discussed in the previous section. We used the intersection of these regions with NSDGeneral (the one we used for training our decoding system), each time creating a synthetic fMRI pattern where the ROI was active (signal set to 1) and the rest of the brain inactive (signal set to 0). From this synthetic pattern, our system could then generate predicted latent variables, and directly reconstruct an equivalent visual scene, corresponding to the ``ROI-optimal'' image. Surprisingly, this simple and deterministic approach, inspired by the analysis in Ozcelik et al~\cite{ozcelik2022reconstruction}, still gives plausible results. Since the synthetic fMRI patterns can be considered out-of-distribution (because there are no similar patterns in the training set), we re-normalized the generated latent variables to give them a similar euclidean norm to the training samples. This procedure helped the diffusion model to generate meaningful images that are shown in Figure~\ref{fig:roi_analysis}.

Upon inspecting the generated ``ROI-optimal'' images for visual ROIs, we see that V1 produces high-contrast scenes with very detailed textures extending to the visual periphery, such as trees and foliage in a park with numerous small human or animal figures. V2 is similar (especially for Sub1 and Sub5, which also display humans in a luxuriant garden environment), but with slightly broader elements and less peripheral detail (e.g. trays filled with various foods in Sub2, Sub7, and in the subject-average). Continuing along the same trend, V3 and V4 produce larger objects compared to the earlier regions, with repeating patterns and global structure. V4 especially generates colorful, high-contrast objects resembling toys on a bright background.

The ROI-optimal images for functionally defined high-level ROIs are even easier to interpret, as they tend to coincide with each region's known category preference. For instance, the model generated multiple face images for the Face-ROI, including humans and sometimes even animal faces (e.g. dogs in Sub5 and in the subject-average). For the Word-ROI, the model generated characters and pseudo-words on objects or signs (except for Sub7). Architecturally plausible indoor scene layouts were produced for the Place-ROI. Finally, for the Body-ROI, the reconstructed images show both human and animal body parts like arms and legs engaged in active behavior like sports or running.

\begin{figure}[!htb]
    \centering
    \includegraphics[width=0.5\textwidth]{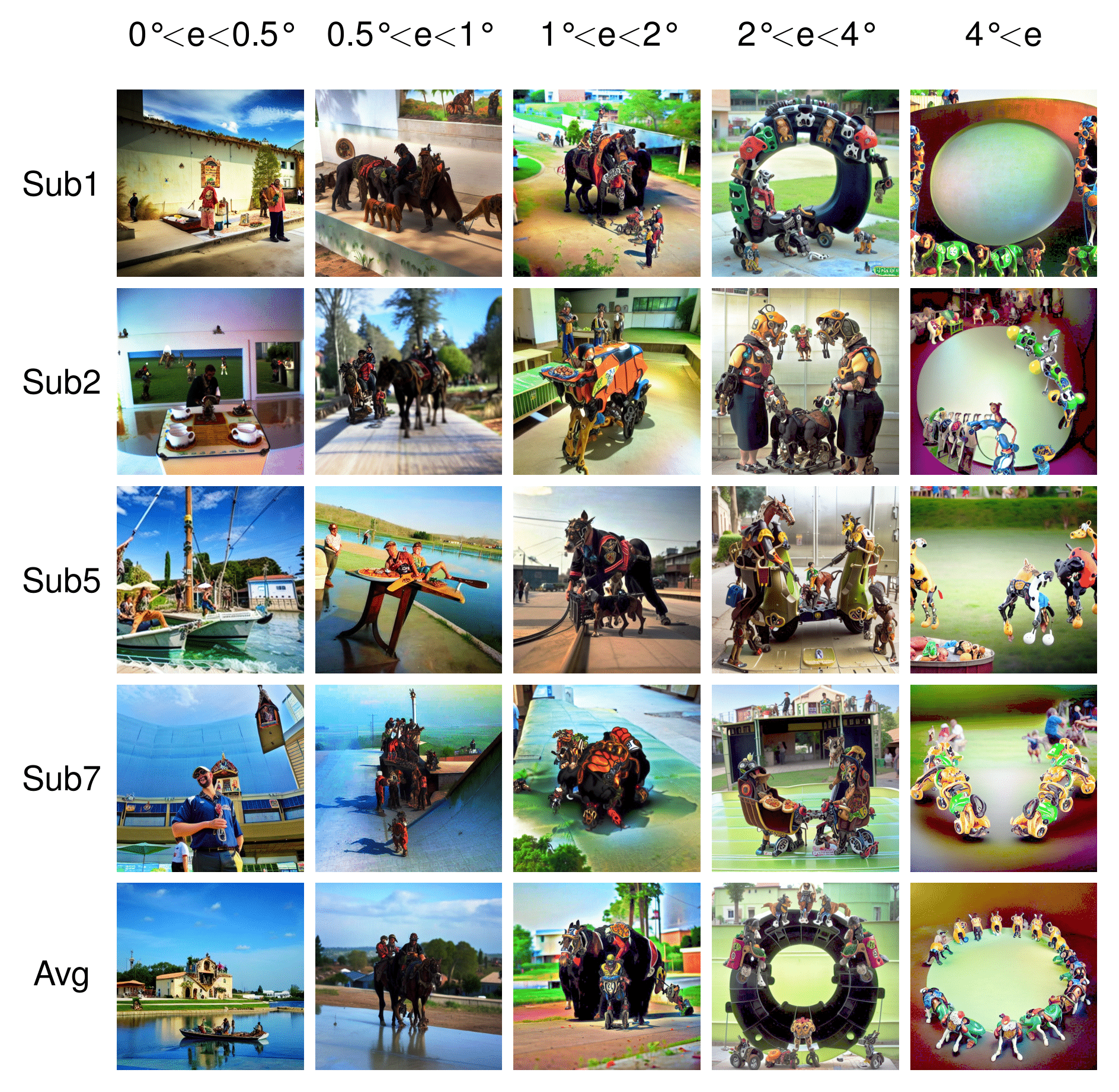}
    \caption{Images reconstructed from synthetic fMRI patterns created by activating regions-of-interests (ROIs) in the visual cortex with different eccentricities. The first 4 rows present individual subjects: Sub1, Sub2, Sub5, and Sub7. The last row is generated by averaging the latent vectors predicted from all 4 subjects. The columns present concentric regions with increasing eccentricity coverage (0°<e<0.5°, 0.5°<e<1°, 1°<e<2°, 2°<e<4°, and 4°<e, where ``e'' stands for eccentricity). Histogram stretching and equalization is applied for visualization purposes.}
    \label{fig:ecc_analysis}
\end{figure}

While these results mainly confirm decades of converging knowledge from the neuroscience literature on neuronal selectivity in the ventral visual pathway, this method allowed us to directly visualize functional properties in vivid detail and high-resolution images. Furthermore, the technique introduced here could easily be extended to study retinotopic or eccentricity-based cortical organization.
As a proof of concept, we also applied our ROI analysis to visual regions defined by different eccentricity preferences. Similar to hierarchical regions in the visual cortex (V1, V2, V3, and hV4) these eccentricity-based regions (0°<e<0.5°, 0.5°<e<1°, 1°<e<2°, 2°<e<4° and 4°<e, where e stands for ``eccentricity'') were also extracted by population receptive field (pRF) experiments. These regions thus reflect the eccentricity preference of the retinotopic cortex, where degrees close to 0° indicate central vision (closer to the fovea) and higher degrees indicate peripheral vision. The corresponding results are shown in Figure~\ref{fig:ecc_analysis}. It is difficult to see a clear pattern for eccentricities between 0° and 0.5° (0°<e<0.5°), as the corresponding portion of the image might be too small to be considered meaningful for the model. A noticeable aspect, however, is that all images for that ROI have detailed and high-contrast objects in the center (though there are also objects in the periphery). For eccentricities between 0.5° and 1° (0.5°<e<1°), and between 1° and 2° (1°<e<2°), we begin to see larger objects (e.g. humans, animals, blobs) at the center of the images. When we reach eccentricities between 2 and 4° (2°<e<4°) and beyond (4°<e), we start to see these objects (or animals, humans, and blobs) move towards the periphery, while the center of the images is mostly empty. These results highlight two important findings: first, the latent representations used by Brain-Diffuser (combining latent features from VDVAE, CLIP-Vision, and CLIP-Text) are precise enough to convey information about the spatial localization of objects in the image; second, we see that Brain-Diffuser managed to learn the eccentricity-based retinotopic organization of the cortex from these representations.

\section{Discussion}

In this study, we designed a two-stage framework (Brain-Diffuser) that reconstructs images from fMRI patterns using generative models based on latent diffusion. In the first stage, we used the VDVAE model to generate ``initial guess'' reconstructions focusing on low-level details. Then in the second stage, we used the image-to-image pipeline of the Versatile Diffusion model, starting from this initial guess, to generate final reconstructions via diffusion, guided by both predicted CLIP-Vision and CLIP-Text features. As we relied on pre-trained and publicly available models for image generation (VDVAE, Versatile Diffusion) and multimodal feature extraction (CLIP), our method only required training ridge regression models from multivoxel brain activity to the relevant model latent spaces (Figures~\ref{fig:vdvae} and ~\ref{fig:vd}).

We analyzed the results both qualitatively (Figure~\ref{fig:multi_subject}) and quantitatively (Table~\ref{tab:method_cmp}) We observed that reconstructed scene images generated by Brain-Diffuser, although not perfectly identical to groundtruth images, preserve most of the layout and semantic information. They also appear more naturalistic compared to reconstructions from earlier studies (Figure~\ref{fig:method_comparison}). When evaluated quantitatively, we saw that Brain-Diffuser outperforms previous models in both high-level and low-level metrics. After advancing the state-of-the-art in image generation applications~\cite{rombach2022high,ramesh2022hierarchical,nichol2022glide,saharia2022photorealistic,xu2022versatile}, it appears that latent diffusion models can also be used to improve the state-of-the-art in fMRI-based image reconstruction. 

Although latent diffusion models are very recent~\cite{rombach2022high}, we noted at least two competing studies that used LDMs for fMRI-based image reconstruction. Chen et al.~\cite{chen2023seeing} proposed MinD-Vis, a method based on an LDM conditioned on image category labels (rather than text captions) to reconstruct images from the Kamitani dataset. As mentioned above, this is a less challenging, single-object-centered dataset; thus, their results cannot be directly compared with ours, obtained using the richer and more complex NSD dataset. Takagi et al.~\cite{takagi2023high}, on the other hand, used NSD and were thus included in our quantitative comparisons. There are multiple possible reasons why our model performed better than theirs, on both low-level and high-level metrics. Beyond the use of distinct pretrained LDMs (Stable Diffusion~\cite{rombach2022high} for Takagi et al. vs. Versatile Diffusion~\cite{xu2022versatile} in our study), our framework contains several improvements such as the use of VDVAE reconstructions for low-level details (Figure~\ref{fig:vdvae}) and the dual conditioning on CLIP-Vision and CLIP-Text features (Figure~\ref{fig:vd}), which together resulted in better qualitative and quantitative results.

There are several ways in which this work may be pursued in the future. As deep generative models will likely continue to improve at a breakneck pace, it is probable that there will soon come models better suited for complex scene reconstruction from fMRI signals. Of course, among a pool of many generative models, it may not be a trivial task to select the most appropriate ones and to experiment on them, and adapt them for brain decoding and image reconstruction. If future generative models reach a ceiling in their ability to linearly explain brain activity, we may need to look for better alternatives than just doing ridge regression between fMRI patterns and latent variables. These alternatives (non-linear regressions, deep hierarchical networks), however, may require larger training datasets to learn the correspondence between fMRI patterns and visual features than ridge regression. We may also see more accurate movie reconstruction studies that process temporal patterns together with spatial ones on movie-fMRI datasets~\cite{wang2022reconstructing,kupershmidt2022penny}.  Besides improving the reconstruction quality, future work could also design novel experiments and analyses on the NSD dataset using generative models. For instance, in this study, we have shown that we can use generative models to reveal the ``optimal'' stimulus for anatomically, functionally, or retinotopically-defined ROIs, by analyzing the reconstructions of synthetic fMRI patterns created from the corresponding ROI masks. This approach could easily be extended to probe less well-known regions of the visual cortex, to help settle theoretical arguments about distinct sub-regions of (e.g.) the face processing network, or to render images for arbitrary combinations of ROIs (e.g., what image would optimally activate V1, V4, and the face-ROI, but not V2 or the Body-ROI). Important advances in this direction were made using an iterative optimization method by Gu et al (2022)~\cite{gu2022neurogen} Directly passing synthetic fMRI patterns to the image reconstruction pipeline, however, is computationally advantageous, which may prove important when there are numerous combinations of sub-regions to be tried. Similar ``virtual experiments'' in this framework could help us address outstanding questions in neuroscience, and understand the organization of sensory and semantic knowledge in the brain.

\section*{Data availability}
The information on accessing the NSD dataset that is analyzed during the current study is available in the Natural Scenes Dataset repository, http://naturalscenesdataset.org/

\bibliography{refs}

\begin{thebibliography}{10}
\urlstyle{rm}
\expandafter\ifx\csname url\endcsname\relax
  \def\url#1{\texttt{#1}}\fi
\expandafter\ifx\csname urlprefix\endcsname\relax\def\urlprefix{URL }\fi
\expandafter\ifx\csname doiprefix\endcsname\relax\def\doiprefix{DOI: }\fi
\providecommand{\bibinfo}[2]{#2}
\providecommand{\eprint}[2][]{\url{#2}}

\bibitem{thirion2006inverse}
\bibinfo{author}{Thirion, B.} \emph{et~al.}
\newblock \bibinfo{journal}{\bibinfo{title}{Inverse retinotopy: inferring the
  visual content of images from brain activation patterns}}.
\newblock {\emph{\JournalTitle{Neuroimage}}} \textbf{\bibinfo{volume}{33}},
  \bibinfo{pages}{1104--1116} (\bibinfo{year}{2006}).

\bibitem{kamitani2005decoding}
\bibinfo{author}{Kamitani, Y.} \& \bibinfo{author}{Tong, F.}
\newblock \bibinfo{journal}{\bibinfo{title}{Decoding the visual and subjective
  contents of the human brain}}.
\newblock {\emph{\JournalTitle{Nature neuroscience}}}
  \textbf{\bibinfo{volume}{8}}, \bibinfo{pages}{679--685}
  (\bibinfo{year}{2005}).

\bibitem{haynes2005predicting}
\bibinfo{author}{Haynes, J.-D.} \& \bibinfo{author}{Rees, G.}
\newblock \bibinfo{journal}{\bibinfo{title}{Predicting the orientation of
  invisible stimuli from activity in human primary visual cortex}}.
\newblock {\emph{\JournalTitle{Nature neuroscience}}}
  \textbf{\bibinfo{volume}{8}}, \bibinfo{pages}{686--691}
  (\bibinfo{year}{2005}).

\bibitem{haxby2001distributed}
\bibinfo{author}{Haxby, J.~V.} \emph{et~al.}
\newblock \bibinfo{journal}{\bibinfo{title}{Distributed and overlapping
  representations of faces and objects in ventral temporal cortex}}.
\newblock {\emph{\JournalTitle{Science}}} \textbf{\bibinfo{volume}{293}},
  \bibinfo{pages}{2425--2430} (\bibinfo{year}{2001}).

\bibitem{cox2003functional}
\bibinfo{author}{Cox, D.~D.} \& \bibinfo{author}{Savoy, R.~L.}
\newblock \bibinfo{journal}{\bibinfo{title}{Functional magnetic resonance
  imaging (fmri)“brain reading”: detecting and classifying distributed
  patterns of fmri activity in human visual cortex}}.
\newblock {\emph{\JournalTitle{Neuroimage}}} \textbf{\bibinfo{volume}{19}},
  \bibinfo{pages}{261--270} (\bibinfo{year}{2003}).

\bibitem{kay2008identifying}
\bibinfo{author}{Kay, K.~N.}, \bibinfo{author}{Naselaris, T.},
  \bibinfo{author}{Prenger, R.~J.} \& \bibinfo{author}{Gallant, J.~L.}
\newblock \bibinfo{journal}{\bibinfo{title}{Identifying natural images from
  human brain activity}}.
\newblock {\emph{\JournalTitle{Nature}}} \textbf{\bibinfo{volume}{452}},
  \bibinfo{pages}{352--355} (\bibinfo{year}{2008}).

\bibitem{miyawaki2008visual}
\bibinfo{author}{Miyawaki, Y.} \emph{et~al.}
\newblock \bibinfo{journal}{\bibinfo{title}{Visual image reconstruction from
  human brain activity using a combination of multiscale local image
  decoders}}.
\newblock {\emph{\JournalTitle{Neuron}}} \textbf{\bibinfo{volume}{60}},
  \bibinfo{pages}{915--929} (\bibinfo{year}{2008}).

\bibitem{vanrullen2019reconstructing}
\bibinfo{author}{VanRullen, R.} \& \bibinfo{author}{Reddy, L.}
\newblock \bibinfo{journal}{\bibinfo{title}{Reconstructing faces from fmri
  patterns using deep generative neural networks}}.
\newblock {\emph{\JournalTitle{Communications biology}}}
  \textbf{\bibinfo{volume}{2}}, \bibinfo{pages}{1--10} (\bibinfo{year}{2019}).

\bibitem{dado2022hyperrealistic}
\bibinfo{author}{Dado, T.} \emph{et~al.}
\newblock \bibinfo{journal}{\bibinfo{title}{Hyperrealistic neural decoding for
  reconstructing faces from fmri activations via the gan latent space}}.
\newblock {\emph{\JournalTitle{Scientific reports}}}
  \textbf{\bibinfo{volume}{12}}, \bibinfo{pages}{141} (\bibinfo{year}{2022}).

\bibitem{shen2019deep}
\bibinfo{author}{Shen, G.}, \bibinfo{author}{Horikawa, T.},
  \bibinfo{author}{Majima, K.} \& \bibinfo{author}{Kamitani, Y.}
\newblock \bibinfo{journal}{\bibinfo{title}{Deep image reconstruction from
  human brain activity}}.
\newblock {\emph{\JournalTitle{PLoS computational biology}}}
  \textbf{\bibinfo{volume}{15}}, \bibinfo{pages}{e1006633}
  (\bibinfo{year}{2019}).

\bibitem{allen2022massive}
\bibinfo{author}{Allen, E.~J.} \emph{et~al.}
\newblock \bibinfo{journal}{\bibinfo{title}{A massive 7t fmri dataset to bridge
  cognitive neuroscience and artificial intelligence}}.
\newblock {\emph{\JournalTitle{Nature neuroscience}}}
  \textbf{\bibinfo{volume}{25}}, \bibinfo{pages}{116--126}
  (\bibinfo{year}{2022}).

\bibitem{lin2022mind}
\bibinfo{author}{Lin, S.}, \bibinfo{author}{Sprague, T.~C.} \&
  \bibinfo{author}{Singh, A.}
\newblock \bibinfo{title}{Mind reader: Reconstructing complex images from brain
  activities}.
\newblock In \bibinfo{editor}{Oh, A.~H.}, \bibinfo{editor}{Agarwal, A.},
  \bibinfo{editor}{Belgrave, D.} \& \bibinfo{editor}{Cho, K.} (eds.)
  \emph{\bibinfo{booktitle}{Advances in Neural Information Processing Systems}}
  (\bibinfo{year}{2022}).

\bibitem{horikawa2017generic}
\bibinfo{author}{Horikawa, T.} \& \bibinfo{author}{Kamitani, Y.}
\newblock \bibinfo{journal}{\bibinfo{title}{Generic decoding of seen and
  imagined objects using hierarchical visual features}}.
\newblock {\emph{\JournalTitle{Nature communications}}}
  \textbf{\bibinfo{volume}{8}}, \bibinfo{pages}{1--15} (\bibinfo{year}{2017}).

\bibitem{deng2009imagenet}
\bibinfo{author}{Deng, J.} \emph{et~al.}
\newblock \bibinfo{title}{Imagenet: A large-scale hierarchical image database}.
\newblock In \emph{\bibinfo{booktitle}{2009 IEEE conference on computer vision
  and pattern recognition}}, \bibinfo{pages}{248--255}
  (\bibinfo{organization}{Ieee}, \bibinfo{year}{2009}).

\bibitem{beliy2019voxels}
\bibinfo{author}{Beliy, R.} \emph{et~al.}
\newblock \bibinfo{journal}{\bibinfo{title}{From voxels to pixels and back:
  Self-supervision in natural-image reconstruction from fmri}}.
\newblock {\emph{\JournalTitle{Advances in Neural Information Processing
  Systems}}} \textbf{\bibinfo{volume}{32}} (\bibinfo{year}{2019}).

\bibitem{gaziv2020self}
\bibinfo{author}{Gaziv, G.} \emph{et~al.}
\newblock \bibinfo{journal}{\bibinfo{title}{Self-supervised natural image
  reconstruction and large-scale semantic classification from brain activity}}.
\newblock {\emph{\JournalTitle{NeuroImage}}} \textbf{\bibinfo{volume}{254}},
  \bibinfo{pages}{119121} (\bibinfo{year}{2022}).

\bibitem{mozafari2020reconstructing}
\bibinfo{author}{Mozafari, M.}, \bibinfo{author}{Reddy, L.} \&
  \bibinfo{author}{VanRullen, R.}
\newblock \bibinfo{title}{Reconstructing natural scenes from fmri patterns
  using bigbigan}.
\newblock In \emph{\bibinfo{booktitle}{2020 International joint conference on
  neural networks (IJCNN)}}, \bibinfo{pages}{1--8}
  (\bibinfo{organization}{IEEE}, \bibinfo{year}{2020}).

\bibitem{ren2021reconstructing}
\bibinfo{author}{Ren, Z.} \emph{et~al.}
\newblock \bibinfo{journal}{\bibinfo{title}{Reconstructing seen image from
  brain activity by visually-guided cognitive representation and adversarial
  learning}}.
\newblock {\emph{\JournalTitle{NeuroImage}}} \textbf{\bibinfo{volume}{228}},
  \bibinfo{pages}{117602} (\bibinfo{year}{2021}).

\bibitem{ozcelik2022reconstruction}
\bibinfo{author}{Ozcelik, F.}, \bibinfo{author}{Choksi, B.},
  \bibinfo{author}{Mozafari, M.}, \bibinfo{author}{Reddy, L.} \&
  \bibinfo{author}{VanRullen, R.}
\newblock \bibinfo{title}{Reconstruction of perceived images from fmri patterns
  and semantic brain exploration using instance-conditioned gans}.
\newblock In \emph{\bibinfo{booktitle}{2022 International Joint Conference on
  Neural Networks (IJCNN)}}, \bibinfo{pages}{1--8}
  (\bibinfo{organization}{IEEE}, \bibinfo{year}{2022}).

\bibitem{chen2023seeing}
\bibinfo{author}{Chen, Z.}, \bibinfo{author}{Qing, J.}, \bibinfo{author}{Xiang,
  T.}, \bibinfo{author}{Yue, W.~L.} \& \bibinfo{author}{Zhou, J.~H.}
\newblock \bibinfo{title}{Seeing beyond the brain: Conditional diffusion model
  with sparse masked modeling for vision decoding}.
\newblock In \emph{\bibinfo{booktitle}{Proceedings of the IEEE/CVF Conference
  on Computer Vision and Pattern Recognition}}, \bibinfo{pages}{22710--22720}
  (\bibinfo{year}{2023}).

\bibitem{lin2014microsoft}
\bibinfo{author}{Lin, T.-Y.} \emph{et~al.}
\newblock \bibinfo{title}{Microsoft coco: Common objects in context}.
\newblock In \emph{\bibinfo{booktitle}{Computer Vision--ECCV 2014: 13th
  European Conference, Zurich, Switzerland, September 6-12, 2014, Proceedings,
  Part V 13}}, \bibinfo{pages}{740--755} (\bibinfo{organization}{Springer},
  \bibinfo{year}{2014}).

\bibitem{takagi2023high}
\bibinfo{author}{Takagi, Y.} \& \bibinfo{author}{Nishimoto, S.}
\newblock \bibinfo{title}{High-resolution image reconstruction with latent
  diffusion models from human brain activity}.
\newblock In \emph{\bibinfo{booktitle}{Proceedings of the IEEE/CVF Conference
  on Computer Vision and Pattern Recognition}}, \bibinfo{pages}{14453--14463}
  (\bibinfo{year}{2023}).

\bibitem{gu2023decoding}
\bibinfo{author}{Gu, Z.}, \bibinfo{author}{Jamison, K.},
  \bibinfo{author}{Kuceyeski, A.} \& \bibinfo{author}{Sabuncu, M.~R.}
\newblock \bibinfo{title}{Decoding natural image stimuli from f{MRI} data with
  a surface-based convolutional network}.
\newblock In \emph{\bibinfo{booktitle}{Medical Imaging with Deep Learning}}
  (\bibinfo{year}{2023}).

\bibitem{rombach2022high}
\bibinfo{author}{Rombach, R.}, \bibinfo{author}{Blattmann, A.},
  \bibinfo{author}{Lorenz, D.}, \bibinfo{author}{Esser, P.} \&
  \bibinfo{author}{Ommer, B.}
\newblock \bibinfo{title}{High-resolution image synthesis with latent diffusion
  models}.
\newblock In \emph{\bibinfo{booktitle}{Proceedings of the IEEE/CVF Conference
  on Computer Vision and Pattern Recognition}}, \bibinfo{pages}{10684--10695}
  (\bibinfo{year}{2022}).

\bibitem{ramesh2022hierarchical}
\bibinfo{author}{Ramesh, A.}, \bibinfo{author}{Dhariwal, P.},
  \bibinfo{author}{Nichol, A.}, \bibinfo{author}{Chu, C.} \&
  \bibinfo{author}{Chen, M.}
\newblock \bibinfo{journal}{\bibinfo{title}{Hierarchical text-conditional image
  generation with clip latents}}.
\newblock {\emph{\JournalTitle{arXiv preprint arXiv:2204.06125}}}
  (\bibinfo{year}{2022}).

\bibitem{nichol2022glide}
\bibinfo{author}{Nichol, A.~Q.} \emph{et~al.}
\newblock \bibinfo{title}{Glide: Towards photorealistic image generation and
  editing with text-guided diffusion models}.
\newblock In \emph{\bibinfo{booktitle}{International Conference on Machine
  Learning}}, \bibinfo{pages}{16784--16804} (\bibinfo{organization}{PMLR},
  \bibinfo{year}{2022}).

\bibitem{saharia2022photorealistic}
\bibinfo{author}{Saharia, C.} \emph{et~al.}
\newblock \bibinfo{title}{Photorealistic text-to-image diffusion models with
  deep language understanding}.
\newblock In \bibinfo{editor}{Oh, A.~H.}, \bibinfo{editor}{Agarwal, A.},
  \bibinfo{editor}{Belgrave, D.} \& \bibinfo{editor}{Cho, K.} (eds.)
  \emph{\bibinfo{booktitle}{Advances in Neural Information Processing Systems}}
  (\bibinfo{year}{2022}).

\bibitem{xu2022versatile}
\bibinfo{author}{Xu, X.}, \bibinfo{author}{Wang, Z.}, \bibinfo{author}{Zhang,
  E.}, \bibinfo{author}{Wang, K.} \& \bibinfo{author}{Shi, H.}
\newblock \bibinfo{journal}{\bibinfo{title}{Versatile diffusion: Text, images
  and variations all in one diffusion model}}.
\newblock {\emph{\JournalTitle{arXiv preprint arXiv:2211.08332}}}
  (\bibinfo{year}{2022}).

\bibitem{radford2021learning}
\bibinfo{author}{Radford, A.} \emph{et~al.}
\newblock \bibinfo{title}{Learning transferable visual models from natural
  language supervision}.
\newblock In \emph{\bibinfo{booktitle}{International conference on machine
  learning}}, \bibinfo{pages}{8748--8763} (\bibinfo{organization}{PMLR},
  \bibinfo{year}{2021}).

\bibitem{child2021very}
\bibinfo{author}{Child, R.}
\newblock \bibinfo{title}{Very deep {\{}vae{\}}s generalize autoregressive
  models and can outperform them on images}.
\newblock In \emph{\bibinfo{booktitle}{International Conference on Learning
  Representations}} (\bibinfo{year}{2021}).

\bibitem{kingma2013auto}
\bibinfo{author}{Kingma, D.~P.} \& \bibinfo{author}{Welling, M.}
\newblock \bibinfo{journal}{\bibinfo{title}{Auto-encoding variational bayes}}.
\newblock {\emph{\JournalTitle{arXiv preprint arXiv:1312.6114}}}
  (\bibinfo{year}{2013}).

\bibitem{schuhmann2021laion}
\bibinfo{author}{Schuhmann, C.} \emph{et~al.}
\newblock \bibinfo{journal}{\bibinfo{title}{Laion-400m: Open dataset of
  clip-filtered 400 million image-text pairs}}.
\newblock {\emph{\JournalTitle{arXiv preprint arXiv:2111.02114}}}
  (\bibinfo{year}{2021}).

\bibitem{wang2004image}
\bibinfo{author}{Wang, Z.}, \bibinfo{author}{Bovik, A.~C.},
  \bibinfo{author}{Sheikh, H.~R.} \& \bibinfo{author}{Simoncelli, E.~P.}
\newblock \bibinfo{journal}{\bibinfo{title}{Image quality assessment: from
  error visibility to structural similarity}}.
\newblock {\emph{\JournalTitle{IEEE transactions on image processing}}}
  \textbf{\bibinfo{volume}{13}}, \bibinfo{pages}{600--612}
  (\bibinfo{year}{2004}).

\bibitem{krizhevsky2017imagenet}
\bibinfo{author}{Krizhevsky, A.}, \bibinfo{author}{Sutskever, I.} \&
  \bibinfo{author}{Hinton, G.~E.}
\newblock \bibinfo{journal}{\bibinfo{title}{Imagenet classification with deep
  convolutional neural networks}}.
\newblock {\emph{\JournalTitle{Communications of the ACM}}}
  \textbf{\bibinfo{volume}{60}}, \bibinfo{pages}{84--90}
  (\bibinfo{year}{2017}).

\bibitem{szegedy2016rethinking}
\bibinfo{author}{Szegedy, C.}, \bibinfo{author}{Vanhoucke, V.},
  \bibinfo{author}{Ioffe, S.}, \bibinfo{author}{Shlens, J.} \&
  \bibinfo{author}{Wojna, Z.}
\newblock \bibinfo{title}{Rethinking the inception architecture for computer
  vision}.
\newblock In \emph{\bibinfo{booktitle}{Proceedings of the IEEE conference on
  computer vision and pattern recognition}}, \bibinfo{pages}{2818--2826}
  (\bibinfo{year}{2016}).

\bibitem{tan2019efficientnet}
\bibinfo{author}{Tan, M.} \& \bibinfo{author}{Le, Q.}
\newblock \bibinfo{title}{Efficientnet: Rethinking model scaling for
  convolutional neural networks}.
\newblock In \emph{\bibinfo{booktitle}{International conference on machine
  learning}}, \bibinfo{pages}{6105--6114} (\bibinfo{organization}{PMLR},
  \bibinfo{year}{2019}).

\bibitem{caron2020unsupervised}
\bibinfo{author}{Caron, M.} \emph{et~al.}
\newblock \bibinfo{journal}{\bibinfo{title}{Unsupervised learning of visual
  features by contrasting cluster assignments}}.
\newblock {\emph{\JournalTitle{Advances in Neural Information Processing
  Systems}}} \textbf{\bibinfo{volume}{33}}, \bibinfo{pages}{9912--9924}
  (\bibinfo{year}{2020}).

\bibitem{hubel1962receptive}
\bibinfo{author}{Hubel, D.~H.} \& \bibinfo{author}{Wiesel, T.~N.}
\newblock \bibinfo{journal}{\bibinfo{title}{Receptive fields, binocular
  interaction and functional architecture in the cat's visual cortex}}.
\newblock {\emph{\JournalTitle{The Journal of physiology}}}
  \textbf{\bibinfo{volume}{160}}, \bibinfo{pages}{106} (\bibinfo{year}{1962}).

\bibitem{gross1972visual}
\bibinfo{author}{Gross, C.~G.}, \bibinfo{author}{Rocha-Miranda, C.~d.} \&
  \bibinfo{author}{Bender, D.}
\newblock \bibinfo{journal}{\bibinfo{title}{Visual properties of neurons in
  inferotemporal cortex of the macaque.}}
\newblock {\emph{\JournalTitle{Journal of neurophysiology}}}
  \textbf{\bibinfo{volume}{35}}, \bibinfo{pages}{96--111}
  (\bibinfo{year}{1972}).

\bibitem{perrett1982visual}
\bibinfo{author}{Perrett, D.}, \bibinfo{author}{Rolls, E.} \&
  \bibinfo{author}{Caan, W.}
\newblock \bibinfo{journal}{\bibinfo{title}{Visual neurones responsive to faces
  in the monkey temporal cortex}}.
\newblock {\emph{\JournalTitle{Experimental brain research}}}
  \textbf{\bibinfo{volume}{47}}, \bibinfo{pages}{329--342}
  (\bibinfo{year}{1982}).

\bibitem{gallant1993selectivity}
\bibinfo{author}{Gallant, J.~L.}, \bibinfo{author}{Braun, J.} \&
  \bibinfo{author}{Van~Essen, D.~C.}
\newblock \bibinfo{journal}{\bibinfo{title}{Selectivity for polar, hyperbolic,
  and cartesian gratings in macaque visual cortex}}.
\newblock {\emph{\JournalTitle{Science}}} \textbf{\bibinfo{volume}{259}},
  \bibinfo{pages}{100--103} (\bibinfo{year}{1993}).

\bibitem{van1994neural}
\bibinfo{author}{Van~Essen, D.~C.} \& \bibinfo{author}{Gallant, J.~L.}
\newblock \bibinfo{journal}{\bibinfo{title}{Neural mechanisms of form and
  motion processing in the primate visual system}}.
\newblock {\emph{\JournalTitle{Neuron}}} \textbf{\bibinfo{volume}{13}},
  \bibinfo{pages}{1--10} (\bibinfo{year}{1994}).

\bibitem{kanwisher1997fusiform}
\bibinfo{author}{Kanwisher, N.}, \bibinfo{author}{McDermott, J.} \&
  \bibinfo{author}{Chun, M.~M.}
\newblock \bibinfo{journal}{\bibinfo{title}{The fusiform face area: a module in
  human extrastriate cortex specialized for face perception}}.
\newblock {\emph{\JournalTitle{Journal of neuroscience}}}
  \textbf{\bibinfo{volume}{17}}, \bibinfo{pages}{4302--4311}
  (\bibinfo{year}{1997}).

\bibitem{epstein1998cortical}
\bibinfo{author}{Epstein, R.} \& \bibinfo{author}{Kanwisher, N.}
\newblock \bibinfo{journal}{\bibinfo{title}{A cortical representation of the
  local visual environment}}.
\newblock {\emph{\JournalTitle{Nature}}} \textbf{\bibinfo{volume}{392}},
  \bibinfo{pages}{598--601} (\bibinfo{year}{1998}).

\bibitem{orban2004comparative}
\bibinfo{author}{Orban, G.~A.}, \bibinfo{author}{Van~Essen, D.} \&
  \bibinfo{author}{Vanduffel, W.}
\newblock \bibinfo{journal}{\bibinfo{title}{Comparative mapping of higher
  visual areas in monkeys and humans}}.
\newblock {\emph{\JournalTitle{Trends in cognitive sciences}}}
  \textbf{\bibinfo{volume}{8}}, \bibinfo{pages}{315--324}
  (\bibinfo{year}{2004}).

\bibitem{bashivan2019neural}
\bibinfo{author}{Bashivan, P.}, \bibinfo{author}{Kar, K.} \&
  \bibinfo{author}{DiCarlo, J.~J.}
\newblock \bibinfo{journal}{\bibinfo{title}{Neural population control via deep
  image synthesis}}.
\newblock {\emph{\JournalTitle{Science}}} \textbf{\bibinfo{volume}{364}},
  \bibinfo{pages}{eaav9436} (\bibinfo{year}{2019}).

\bibitem{ratan2021computational}
\bibinfo{author}{Ratan~Murty, N.~A.}, \bibinfo{author}{Bashivan, P.},
  \bibinfo{author}{Abate, A.}, \bibinfo{author}{DiCarlo, J.~J.} \&
  \bibinfo{author}{Kanwisher, N.}
\newblock \bibinfo{journal}{\bibinfo{title}{Computational models of
  category-selective brain regions enable high-throughput tests of
  selectivity}}.
\newblock {\emph{\JournalTitle{Nature communications}}}
  \textbf{\bibinfo{volume}{12}}, \bibinfo{pages}{5540} (\bibinfo{year}{2021}).

\bibitem{gu2022neurogen}
\bibinfo{author}{Gu, Z.} \emph{et~al.}
\newblock \bibinfo{journal}{\bibinfo{title}{Neurogen: activation optimized
  image synthesis for discovery neuroscience}}.
\newblock {\emph{\JournalTitle{NeuroImage}}} \textbf{\bibinfo{volume}{247}},
  \bibinfo{pages}{118812} (\bibinfo{year}{2022}).

\bibitem{mueller2023macaques}
\bibinfo{author}{Mueller, K.~N.}, \bibinfo{author}{Carter, M.~C.},
  \bibinfo{author}{Kansupada, J.~A.} \& \bibinfo{author}{Ponce, C.~R.}
\newblock \bibinfo{journal}{\bibinfo{title}{Macaques recognize features in
  synthetic images derived from ventral stream neurons}}.
\newblock {\emph{\JournalTitle{Proceedings of the National Academy of
  Sciences}}} \textbf{\bibinfo{volume}{120}}, \bibinfo{pages}{e2213034120}
  (\bibinfo{year}{2023}).

\bibitem{wang2022reconstructing}
\bibinfo{author}{Wang, C.} \emph{et~al.}
\newblock \bibinfo{journal}{\bibinfo{title}{Reconstructing rapid natural vision
  with fmri-conditional video generative adversarial network}}.
\newblock {\emph{\JournalTitle{Cerebral Cortex}}}
  \textbf{\bibinfo{volume}{32}}, \bibinfo{pages}{4502--4511}
  (\bibinfo{year}{2022}).

\bibitem{kupershmidt2022penny}
\bibinfo{author}{Kupershmidt, G.}, \bibinfo{author}{Beliy, R.},
  \bibinfo{author}{Gaziv, G.} \& \bibinfo{author}{Irani, M.}
\newblock \bibinfo{journal}{\bibinfo{title}{A penny for your (visual) thoughts:
  Self-supervised reconstruction of natural movies from brain activity}}.
\newblock {\emph{\JournalTitle{arXiv preprint arXiv:2206.03544}}}
  (\bibinfo{year}{2022}).

\end{thebibliography}

\section*{Acknowledgements}
This work was funded by the Agence Nationale de la Recherche ANR grants AI-REPS ANR-18-CE37-0007-01 and ANITI ANR-19-PI3A-0004.

\section*{Author contributions}
F.O. and R.V. conceptualized the idea. F.O. analyzed the data, designed the framework, and generated the results. F.O. and R.V. analyzed and interpreted the results. F.O. wrote the original draft. F.O. and R.V. reviewed and edited the draft. F.O. and R.V. prepared the figures. All authors reviewed the manuscript.

\section*{Competing interests}
The authors declare no competing interests.
\end{document}